\pdfoutput=1

\documentclass[11pt,a4paper]{article}
\usepackage{acl}
\usepackage{amsfonts}
\usepackage{amsmath}
\usepackage{amssymb}
\usepackage{bm}
\usepackage{booktabs}
\usepackage[T5,T1]{fontenc}
\usepackage{footnotehyper}
\usepackage{graphicx}
\usepackage{multirow}
\usepackage{caption}
\usepackage{subcaption}
\usepackage{svg}
\usepackage{tabularx}
\usepackage{times}
\usepackage{latexsym}

\usepackage{graphicx}
\usepackage{algorithm}
\usepackage{algpseudocode}
\usepackage{euflag}
\algtext*{EndWhile}
\algtext*{EndIf}

\usepackage{anyfontsize}

\DeclareTextSymbolDefault{\ohorn}{T5}
\DeclareTextSymbolDefault{\uhorn}{T5}
\DeclareUnicodeCharacter{0300}{`}

\usepackage{microtype}

\usepackage{todonotes}
\makeatletter
\newcommand*\iftodonotes{\if@todonotes@disabled\expandafter\@secondoftwo\else\expandafter\@firstoftwo\fi}
\makeatother

\definecolor{edolime}{rgb}{0.9,1,0.3}

\newcommand{\method}[1]{{\textsc{#1}}}

\title{Composable Sparse Fine-Tuning for Cross-Lingual Transfer}

\author{Alan Ansell$^1$~~~Edoardo Maria Ponti$^{1,2}$~~~Anna Korhonen$^1$~~~Ivan Vuli\'{c}$^1$ \\
        $^1$Language Technology Lab, University of Cambridge\\
        $^2$Mila - Quebec AI Institute and McGill University\\
}

\date{}

\begin{document}
\maketitle
\begin{abstract}
Fine-tuning the entire set of parameters of a large pretrained model has become the mainstream approach for transfer learning. To increase its efficiency and prevent catastrophic forgetting and interference, techniques like adapters and sparse fine-tuning have been developed. Adapters are \textit{modular}, as they can be combined to adapt a model towards different facets of knowledge (e.g., dedicated language and/or task adapters). Sparse fine-tuning is \textit{expressive}, as it controls the behavior of all model components. In this work, we introduce a new fine-tuning method with \textit{both} these desirable properties. In particular, we learn sparse, real-valued masks based on a simple variant of the Lottery Ticket Hypothesis. Task-specific masks are obtained from annotated data in a source language, and language-specific masks from masked language modeling in a target language. Both these masks can then be composed with the pretrained model. Unlike adapter-based fine-tuning, this method neither increases the number of parameters at inference time nor alters the original model architecture. Most importantly, it outperforms adapters in zero-shot cross-lingual transfer by a large margin in a series of multilingual benchmarks, including Universal Dependencies, MasakhaNER, and AmericasNLI. 
Based on an in-depth analysis, we additionally find that sparsity is crucial to prevent both 1) interference between the fine-tunings to be composed and 2) overfitting. We release the code and models at \url{https://github.com/cambridgeltl/composable-sft}.
\end{abstract}

\begin{figure*}[!t]
    \centering
    \includegraphics[width=0.97\textwidth,trim={0 4.5cm  0 3.5cm},clip]{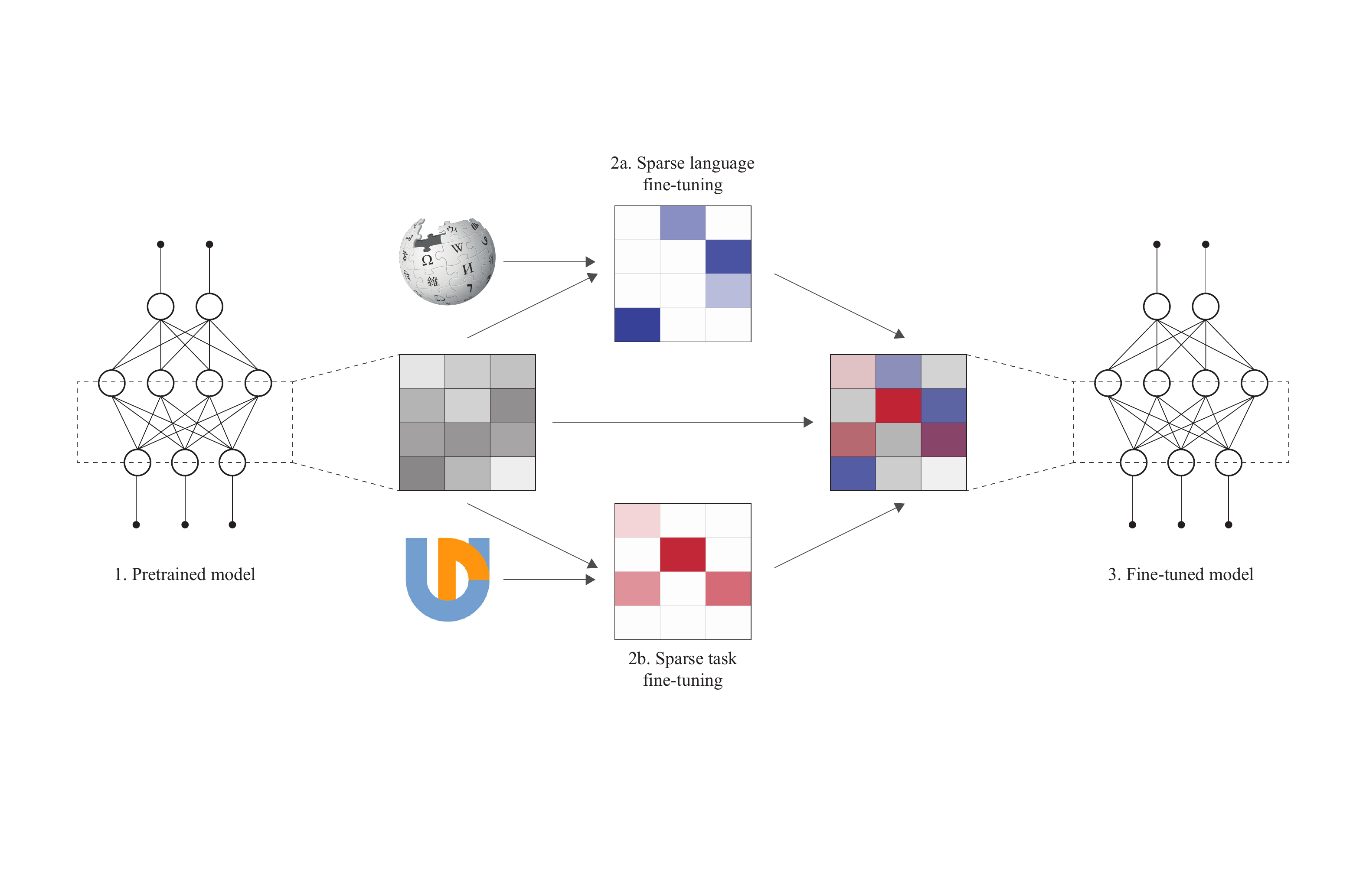}
    \caption{A graphical representation of Lottery Ticket Sparse Fine-Tuning: from the parameters of a pretrained model (gray, left), we generate sparse fine-tunings for task and language knowledge (blue and red, center). Finally, we sum these three components (right) to obtain the adapted/fine-tuned model. Best viewed in color.}
    \label{fig:lt-sft}
\end{figure*}

\section{Introduction}
Fine-tuning of pretrained models \citep[\textit{inter alia}]{howard-ruder-2018-universal,devlin-etal-2019-bert} is arguably the dominant paradigm in NLP at present. Originally, ``fine-tuning'' involved supervised learning of all the parameters of a model pretrained on unlabeled texts. 
However, given the size of Transformer-based architectures, this approach is often time- and resource- inefficient, and may result in catastrophic forgetting and interference \cite{Wang:2020emnlp} during multiple adaptations. 
To overcome these limitations, two main alternatives have emerged: 1) through \textit{adapters}, new parameters can be added to a pretrained model in the form of extra intermediate layers \citep{rebuffi-etal-2017-learning,houlsby-etal-2019-parameter} and fine-tuned while keeping all the pretrained parameters fixed; 2) \textit{sparse} fine-tuning (SFT) of a small subset of pretrained model parameters \citep[\textit{inter alia}]{guo-etal-2021-parameter,zaken-etal-2021-bitfit,xu-etal-2021-raise}.

Adapters have proven especially useful in multilingual NLP \citep{Bapna:2019emnlp,Ustun:2020emnlp,pfeiffer-etal-2020-mad,Vidoni:2020arxiv,pfeiffer-etal-2021-unks,ansell-etal-2021-mad-g} because they exhibit a surprising degree of \textit{modularity}. This ability to disentangle and recombine orthogonal facets of knowledge in original ways \citep{TACL2123,ponti2021inductive} allows for separately learning a task adapter from labeled data in a source language and dedicated language adapters from unlabeled data in the source language and target languages. By stacking these components, it is possible to perform zero-shot cross-lingual transfer. Compared to sequentially fine-tuning the full model on both the task and target language, this yields superior performance and efficiency \citep{pfeiffer-etal-2020-mad}. Notably, achieving coverage over $N_T$ tasks in $N_L$ target languages with the sequential approach requires $N_T N_L$ models to be trained, whereas the modularity of adapters reduces this to $N_T + N_L$.

Meanwhile, the advantage of SFTs over adapters is their \textit{expressivity}: rather than a non-linear transformation of the output of Transformer layers (e.g., using a shallow MLP as with adapters), they can operate directly on a pretrained model's embedding and attention layers. It therefore seems natural to search for a parameter-efficient fine-tuning method that is both modular and expressive.


To this end, we propose Lottery Ticket Sparse Fine-Tuning (LT-SFT), a simple and general-purpose adaptation technique inspired by the Lottery Ticket Hypothesis \citep[LTH;][]{frankle2018the,pmlr-v119-malach20a}, which was originally conceived for pruning large neural networks. In particular, after fine-tuning a pretrained model for a specific task or language, we select the subset of parameters that change the most. Then, we rewind the model to its pretrained initialization (without setting any value to zero, contrary to the original LTH algorithm). By re-tuning again only the selected subset of parameters, we obtain a sparse fine-tuning in the form of a vector of differences with respect to the pretrained model. Multiple SFTs can be \textit{composed} by simply summing them with the pretrained model. We provide a graphical representation of our method in Figure~\ref{fig:lt-sft}.

We benchmark LT-SFT on a series of multilingual datasets, including Universal Dependencies \citep{ud-2.7} for part-of-speech tagging and dependency parsing, MasakhaNER \citep{adelani-etal-2021-masakhaner} for named entity recognition, and AmericasNLI \citep{ebrahimi-etal-2021-americasnli} for natural language inference. We evaluate it in a zero-shot cross-lingual transfer setting on 35 typologically and geographically diverse languages that include both languages seen and unseen during masked language modeling of the pretrained model. The results in all transfer tasks indicate that LT-SFT consistently achieves substantial gains over the current state-of-the-art adapter-based method for cross-lingual transfer, MAD-X \citep{pfeiffer-etal-2020-mad}.


In addition to its superior performance, modularity, and expressivity, LT-SFT offers a series of additional advantages over adapters: 1) the number of parameters remains constant, which prevents the decrease in inference speed observed when adapter layers are added; 2) the neural architecture remains identical to the pretrained model, which makes code development model-independent rather than requiring special modifications for each possible architecture \citep{pfeiffer-etal-2020-adapterhub}. Finally, 3) we empirically demonstrate that the peak in performance for LT-SFT is consistently found with the same percentage of tunable parameters, whereas the best reduction factor for MAD-X is task-dependent. This makes our method more robust to the choice of hyper-parameters.




In addition, 
we find that a high level of sparsity in language and task fine-tunings is beneficial to performance, as this makes overlaps less likely and poses a lower risk of creating interference between the knowledge they contain. Moreover, it makes fine-tunings less prone to overfitting due to their constrained capacity. Thus, sparsity is a fundamental ingredient for achieving
modularity and composability. These properties in turn allow for systematic generalization to new combinations of tasks and languages in a zero-shot fashion.



\section{Background}
\label{sec:background}
To establish a broader context for our research, we first provide a succinct overview of current methods for efficient fine-tuning, such as adapters and SFT. We then recapitulate the Lottery Ticket Hypothesis, upon which our newly proposed method is built.

\vspace{1.5mm}
\noindent \textbf{Adapters and Composition.} 
An \textit{adapter} is a component inserted into a Transformer model with the purpose of specializing it for a particular language, task, domain, or modality \cite{houlsby-etal-2019-parameter}. Previous work in multilingual NLP has mainly adopted the lightweight yet effective adapter variant of \citet{pfeiffer-etal-2021-adapterfusion}. In this setup, only one adapter module, consisting of a successive down-projection and up-projection, is injected per Transformer layer, after the feed-forward sub-layer. The adapter $\text{A}_b$ at the $b$-th Transformer layer performs the following operation:
\begin{align}
    \text{A}_b(\bm{h}_b, \bm{r}_b) = U_b \, \textrm{a}(D_b\bm{h}_b) + \bm{r}_b.
\end{align}
$\bm{h}_b$ and $\bm{r}_b$ are the Transformer hidden state and the residual at layer $b$, respectively. $D_b \in \mathbb{R}^{m \times h}$ and $U_b \in \mathbb{R}^{h \times m}$ are the down- and up-projections, respectively ($h$ being the Transformer's hidden layer size, and $m$ the adapter's dimension), and $\textrm{a}(\cdot)$ is a non-linear activation function. The residual connection $\bm{r}_b$ is the output of the Transformer's feed-forward layer whereas $\bm{h}_b$ is the output of the subsequent layer normalization. During fine-tuning of a pretrained model with adapters, only the adapter parameters $U$ and $D$ are modified while the pretrained model's parameters are kept fixed.

In the MAD-X adapter composition framework for cross-lingual transfer \citep{pfeiffer-etal-2020-mad}, a \textit{language adapter} (LA) for a massively multilingual Transformer (MMT) is learned for each source and target language through masked language modeling (MLM), and a \textit{task adapter} (TA) is learned for each target task, where the LA for the source language is inserted during TA training. At inference time, the task adapter and target language adapter are \textit{composed} by stacking one on top of the other. This adapter composition approach has been shown to be highly effective for cross-lingual transfer \citep{pfeiffer-etal-2020-mad,pfeiffer-etal-2021-unks,ansell-etal-2021-mad-g}, especially for low-resource languages and target languages unseen during MMT pretraining.


\vspace{1.5mm}
\noindent \textbf{Sparse Fine-Tuning.}
We call $F' = F(\cdot ; \bm{\theta} + \bm{\phi})$ a \textit{sparse fine-tuning} (SFT) of a pretrained neural model $F(\cdot ; \bm{\theta})$ if $\bm{\phi}$ is sparse. We sometimes refer to $\bm\phi$ itself as an SFT, or as the SFT's \textit{difference vector}. Previously proposed SFT methods include DiffPruning \citep{guo-etal-2021-parameter}, BitFit \citep{zaken-etal-2021-bitfit} and ChildTuning \citep{xu-etal-2021-raise}. DiffPruning simulates sparsity of the difference vector during training by applying a continuous relaxation of a binary mask to it. BitFit on the other hand allows non-zero differences only for bias parameters. ChildTuning selects a subset of fine-tunable parameters by using Fisher information to measure the relevance of each parameter to the task. These methods have been shown to be competitive with full fine-tuning on GLUE \citep{wang2018glue}, despite the difference vector $\bm{\phi}$ having fewer than 0.5\% non-zero values.

\vspace{1.5mm}
\noindent \textbf{Lottery Ticket Hypothesis.} \citep[LTH;][]{frankle2018the,pmlr-v119-malach20a} states that each neural model contains a sub-network (a ``winning ticket'') that, if trained again in isolation, can match or even exceed the performance of the original model. To achieve this, after a pruning stage where some parameters are zero-masked and frozen according to some criterion (e.g., weight magnitude), the remaining parameters are restored to their original values and then re-tuned. This process of pruning and re-training can be iterated multiple times. 

The LTH has so far been used mostly for model \textit{compression} through network pruning; to our knowledge, we are the first to use it for pretrained model \textit{adaptation}. 

\vspace{1.5mm}
\noindent \textbf{Multi-Source Task Training.}
\citet{ansell-etal-2021-mad-g} showed that training task adapters using data from multiple source languages can result in sizable improvements in downstream zero-shot transfer performance even when the total number of training examples is held constant. In their training setup, each batch consisted of examples from a single, randomly selected source language, the language adapter for which is activated for the duration of the training step.

\section{Methodology}
\label{sec:methodology}

\subsection{Lottery Ticket Sparse Fine-Tuning} \label{sec:lt-sft}

\noindent \textbf{Training.} In this work, we propose Lottery Ticket Sparse Fine-Tuning (LT-SFT). Similar to the Lottery Ticket algorithm of \citet{frankle2018the}, our LT-SFT method consists of two phases:

\vspace{1mm}
\noindent \textit{(Phase 1)} Pretrained model parameters $\bm{\theta}^{(0)}$ are fully fine-tuned on the target language or task data $\mathcal{D}$, yielding $\bm{\theta}^{(1)}$. Parameters are ranked according to some criterion, in our case greatest absolute difference $|\theta_i^{(1)} - \theta_i^{(0)}|$, and the top $K$ are selected for tuning in the next phase: a binary mask $\bm{\mu}$ is set to have 1 in positions corresponding to these parameters, and 0 elsewhere.

\vspace{1mm}
\noindent \textit{(Phase 2)} After resetting the parameters to their original values $\bm{\theta}^{(0)}$, the model is again fine-tuned, but this time only the $K$ selected parameters are trainable whereas the others are kept frozen. In practice, we implement this by passing the \textit{masked} gradient $ \bm{\mu} \odot \nabla_{\bm\theta} \mathcal{L}(F(\cdot;\bm\theta), \mathcal{D})$ (where $\odot$ denotes element-wise multiplication and $\mathcal{L}$ a loss function) to the optimizer at each step. From the resulting fine-tuned parameters $\bm{\theta}^{(2)}$ we can obtain the sparse vector of differences $\bm{\phi} = \bm{\theta}^{(2)} - \bm{\theta}^{(0)}$.

In addition, we experiment with applying a regularization term which discourages parameters from deviating from their pretrained values $\bm{\theta}^{(0)}$. Specifically, we use L1 regularization of the form $J(\bm{\theta}) = \frac{\lambda}{N} \sum_i |\theta_i - \theta_i^{(0)}|$. 

\vspace{1.5mm}
\noindent \textbf{Composition.} 
Although we often use the term ``sparse fine-tuning'' to refer to the difference vector $\bm{\phi}$ itself, an SFT is most accurately conceptualized as a functional which takes as its argument a parameterized function and returns a new function, where some sparse difference vector $\bm{\phi}$ has been added to the original parameter vector. Suppose we have a language SFT $S_L$ and a task SFT $S_T$ defined by
\vspace{-2mm}
\begin{align*}
    &S_L(F(\cdot ; \bm{\theta})) = F(\cdot ; \bm{\theta} + \bm{\phi}_L) \\
    &S_T(F(\cdot ; \bm{\theta})) = F(\cdot ; \bm{\theta} + \bm{\phi}_T).
\end{align*}
Then we have
\vspace{-2mm}
\begin{align*}
    S_L \circ S_T(F(\cdot ; \bm{\theta})) = F(\cdot ; \bm{\theta} + \bm{\phi}_T + \bm{\phi}_L).
\end{align*}

\subsection{Zero-Shot Transfer with LT-SFT} \label{sec:sft-zs-xlt}
We adopt a similar cross-lingual transfer setup to MAD-X \cite[see also \S\ref{sec:background}]{pfeiffer-etal-2020-mad}. We start with an MMT $F$ with pretrained parameters $\bm{\theta}$ learned through masked language modeling on many languages, such as mBERT \citep{devlin-etal-2019-bert} or XLM-R \citep{conneau-etal-2020-unsupervised}.

For each language of interest $l$, we learn a language SFT $\bm{\phi}_L^{(l)}$ through LT-SFT (also with an MLM objective) on text from language $l$.

For each task of interest $t$, we learn a task SFT $\bm{\phi}_T^{(t)}$ through LT-SFT on annotated data from some source language $s$. When learning the task SFT, we first adapt to the source language by applying the language SFT for $s$.\footnote{Adapting to the source language yields substantial improvements in cross-lingual transfer performance with both MAD-X and LT-SFT, with gains of 2-3 points in our preliminary experiments. Paradoxically, our results (see Table~\ref{tab:seen-ud-results}) and results from previous work \citep{pfeiffer-etal-2020-mad,ansell-etal-2021-mad-g} suggest that adapting to high-resource \textit{target} languages at inference time does not give similarly large benefits. We think this phenomenon warrants further investigation.}
The language SFT is removed again after training. That is, we perform LT-SFT on $F(\cdot\ ; \bm{\theta} + \bm{\phi}_L^{(s)})$ to obtain fine-tuned parameter vector $\bm{\theta}'$. We then calculate $\bm{\phi}_T^{(t)} = \bm{\theta}' - (\bm{\theta} + \bm{\phi}_L^{(s)})$. Note that during task training, we also learn a classifier head, which is fully fine-tuned during both phases of LT-SFT adaptation, with the same random initialization applied at the beginning of each phase.

We perform zero-shot adaptation of $F$ to target language $l$ for task $t$ by composing language and task SFTs to obtain $F_{t,l} = F(\cdot\ ; \bm{\theta} + \bm{\phi}_T^{(t)} + \bm{\phi}_L^{(l)})$. On top of this, we stack the classifier head learned for $t$. For a formal algorithm of LT-SFT and the transfer procedure, we refer to Appendix~\ref{sec:ltsftalg}.

\section{Experimental Setup}

\begin{table*}[!t]
    \centering
    \def\arraystretch{0.95}
    \footnotesize
    \resizebox{\linewidth}{!}{\begin{tabular}{m{0.15\textwidth} | m{0.15\textwidth} | m{0.15\textwidth} | m{0.10\textwidth} | m{0.33\textwidth}}
        \toprule
        \textbf{Task} & \textbf{Target Dataset} & \textbf{Source Dataset} & \textbf{MMT} & \textbf{Target Languages} \\
        \midrule
        Part-of-Speech Tagging (POS), Dependency Parsing (DP) & Universal Dependencies 2.7 \citep{ud-2.7}
        & Universal Dependencies 2.7 \citep{ud-2.7} & mBERT & Arabic$^\dagger$, Bambara, Buryat, Cantonese, Chinese$^\dagger$, Erzya, Faroese, Japanese$^\dagger$, Livvi, Maltese, Manx, North Sami, Komi Zyrian, Sanskrit, Upper Sorbian, Uyghur \\
        \midrule
        Named Entity Recognition (NER) & MasakhaNER \citep{adelani-etal-2021-masakhaner} & CoNLL 2003 \citep{tjong-kim-sang-de-meulder-2003-introduction} & mBERT  & Hausa, Igbo, Kinyarwanda, Luganda, Luo, Nigerian-Pidgin, Swahili$^*$, Wolof, Yor\`{u}b\'{a}$^*$ \\
        \midrule
        Natural Language Inference (NLI) & AmericasNLI \citep{ebrahimi-etal-2021-americasnli} & MultiNLI  \citep{williams-etal-2018-broad} & XLM-R  & Aymara, Ash\'{a}ninka, Bribri, Guarani, N\'{a}huatl, Otom\'{i}, Quechua, Rar\'{a}muri, Shipibo-Konibo, Wixarika \\
        \bottomrule
    \end{tabular}}
    \caption{Details of the tasks, datasets, MMTs and languages involved in our zero-shot cross-lingual transfer evaluation. $^*$ denotes low-resource languages seen during MMT pretraining; $^\dagger$ denotes high-resource languages seen during MMT pretraining; all other languages are low-resource and unseen. The source language is always English. Further details of all the language and data sources used are provided in Appendix \ref{sec:languages}.}
    \vspace{-3mm}
    \label{tab:tasks}
\end{table*}

To evaluate our new method extensively, we benchmark its zero-shot cross-lingual performance on four distinct tasks: part-of-speech tagging (POS), dependency parsing (DP), named entity recognition (NER), and natural language inference (NLI). 
Table \ref{tab:tasks} summarizes our experimental setup, including the datasets and languages considered in our experiments. We put emphasis on low-resource languages and languages unseen during MMT pretraining, although we also evaluate on a few high-resource languages. 
In total, we cover a set of 35 typologically and geographically diverse languages, which makes them representative of cross-lingual variation \citep{ponti-etal-2019-modeling,ponti-etal-2020-xcopa}.

\subsection{Baselines and Model Variants}

The main baseline is \method{MAD-X}, the state-of-the-art adapter-based framework for cross-lingual transfer \citep{pfeiffer-etal-2020-mad}. We use the ``MAD-X 2.0'' variant, where the last adapter layers are dropped. \citet{pfeiffer-etal-2021-unks} found that this improved performance, which we could confirm in our preliminary experiments. Since adapters with the configuration used by \citet{pfeiffer-etal-2020-mad} are unavailable for many languages in our evaluation, we train our own for all languages. In Appendix \ref{sec:hub-eval} we also provide an evaluation with comparable language adapters from AdapterHub \citep{pfeiffer-etal-2020-adapterhub} where available.

We also perform experiments with \method{BitFit} \citep{zaken-etal-2021-bitfit} to establish a baseline for an existing SFT technique. In addition to the main \method{LT-SFT} model variant, on POS and DP we test a \method{rand-SFT} variant as an ablation, where the $K$ parameters to be fine-tuned are selected at random rather than based on an informed criterion.

For both LT-SFT and MAD-X, we also evaluate a task adaptation \method{(TA)-only} configuration, where only the task SFT/adapter is applied, without the target language SFT/adapter.

\subsection{Language SFT/Adapter Training Setup}
\noindent \textbf{MLM Training Data.} 
For all languages in our POS and DP evaluation, we perform MLM language SFT/adapter training on Wikipedia corpora. We also use Wikipedia for all languages in our NER evaluation if available. Where this is not the case, we use the Luo News Dataset \citep{adelani-etal-2021-masakhaner} for Luo and the JW300 corpus \citep{agic-vulic-2019-jw300} for Nigerian Pidgin. The main corpora for the languages in our NLI evaluation are those used by the dataset creators to train their baseline models \citep{ebrahimi-etal-2021-americasnli}; however, since the sizes of these parallel corpora are small, we further augment them with data from Wikipedia and the corpora of indigenous Peruvian languages of \citet{bustamante-etal-2020-data} where available. More details on data sources are provided in Appendix~\ref{sec:languages}.

\vspace{1.5mm}
\noindent \textbf{Training Setup and Hyper-parameters.}
For both SFTs and adapters, we train for the lesser of 100 epochs or 100,000 steps of batch size 8 and maximum sequence length 256, subject to an absolute minimum of 30,000 steps since 100 epochs seemed insufficient for some languages with very small corpora. Model checkpoints are evaluated every 1,000 steps (5,000 for high-resource languages) on a held-out set of 5\% of the corpus (1\% for high-resource languages), and the one with the smallest loss is selected at the end of training. We use the AdamW optimizer \citep{loshchilov-hutter-2019-decoupled} with an initial learning rate of 5$e$-5 which is linearly reduced to 0 over the course of training.

Following \citet{pfeiffer-etal-2020-mad}, the reduction factor (i.e., the ratio between model hidden size and adapter size) for the adapter baseline was set to 2 for a total of $\sim$7.6M trainable parameters. For comparability, we set the same number of trainable parameters $K$ for our language LT-SFTs. This results in language SFTs with a sparsity of 4.3\% for mBERT and 2.8\% for XLM-R. Since \method{BitFit} tunes exclusively the bias parameters, its language SFTs have a fixed sparsity of 0.047\% for mBERT and 0.030\% for XLM-R.


Importantly, during language sparse fine-tuning, we decouple the input and output embedding matrices and fix the parameters of the output matrix; otherwise, we find that the vast majority of the $K$ most changed parameters during full fine-tuning belong to the embedding matrix, seemingly due to its proximity to the model output, which damages downstream performance. We also fix the layer normalization parameters; all other parameters are trainable. 
For language adaptation, we apply L1 regularization as described in \S\ref{sec:lt-sft} with $\lambda = 0.1$.
Note that the specified training regime is applied in the same way during both phases of LT-SFT.

For language adapter training in the MAD-X baseline, we use the Pfeiffer configuration \citep{pfeiffer-etal-2021-adapterfusion} with invertible adapters, special additional sub-components designed for adapting to the vocabulary of the target language.

\subsection{Task SFT/Adapter Training Setup}
For POS tagging, DP, and NER,\footnote{MasakhaNER and CoNLL 2003 datasets respectively use the \texttt{DATE} and \texttt{MISC} tags which are not used by the other; we replace these with the \texttt{O} tag at both train and test time.} we train task SFTs/adapters on the datasets indicated in Table~\ref{tab:tasks} for 10 epochs with batch size 8, except during the first phase of LT-SFT training where we train for only 3 epochs.\footnote{This is because full fine-tuning is more prone to overfitting than sparse/adapter fine-tuning. Early stopping somewhat addresses overfitting, but it is insufficient in a cross-lingual setting because the target language performance generally starts to deteriorate faster than the source language performance.} 
Model checkpoints are evaluated on the validation set every 250 steps, and the best checkpoint is taken at the end of training, with the selection metric being accuracy for POS, labeled attachment score for DP, and F1-score for NER. Similarly to language fine-tuning, we use an initial learning rate of 5$e$-5 which is linearly reduced to 0 over the course of training.  For POS and NER we use the standard token-level single-layer multi-class model head. For DP, we use the shallow variant \citep{glavas-vulic-2021-supervised} of the biaffine dependency parser of \citet{dozat-manning-2017-deep}. 

For NLI, we employ the same fine-tuning hyper-parameters as \citet{ebrahimi-etal-2021-americasnli}: 5 epochs with batch size 32, with checkpoint evaluation on the validation set every 625 steps, and an initial learning rate of 2$e$-5. We apply a two-layer multi-class classification head atop the MMT output corresponding to the \texttt{[CLS]} token.

We found that the number of trainable parameters during task adaptation (governed by $K$ for SFTs and reduction factor for adapters) has a large effect on performance: we thus experiment with a range of values. Specifically, we test adapter reduction factors of 32, 16, 8, 4, 2, and 1, and equivalent values of $K$ \footnote{Approximately 442K, 884K, 1.7M, 3.5M, 7.1M, and 14.2M respectively, amounting to sparsity levels of 0.25\%, 0.50\%, 1.0\%, 2.0\%, 4.0\% and 8.0\% for mBERT and 0.16\%, 0.32\%, 0.63\%, 1.3\%, 2.6\% and 5.1\% for XLM-R.} for SFT.

During task adaptation, we always apply the source language adapter following \citet{pfeiffer-etal-2020-mad}, or source language SFT (see \S\ref{sec:sft-zs-xlt}).

\subsection{Multi-Source Training}
\label{ss:exp-ms}
To validate that task LT-SFT training, like task adapter training in prior work \cite{ansell-etal-2021-mad-g}, benefits from the presence of multiple source languages in the training data, and to push the boundaries of zero-shot cross lingual transfer, we perform multi-source training experiments on DP and NLI.

We adopt a similar setup to \citet{ansell-etal-2021-mad-g}: we obtain the training set by concatenating the training data for all source languages. We randomly shuffle the training set and train as in the single-source case, except that each batch is composed of examples from a single source language, whose language SFT is applied during the training step.

We prioritize maximizing performance rather than providing a fair comparison against the single-source case, so unlike \citet{ansell-etal-2021-mad-g}, we use the entirety of the training sets. In derogation of this principle, we set a maximum of 15K examples per language for DP to better balance our sample.

For DP, we train our models on the UD treebanks of 11 diverse high-resource languages. 
For NLI, we train on MultiNLI \citep{williams-etal-2018-broad} plus the data for all 14 non-English languages in the XNLI dataset \citep{conneau-etal-2018-xnli}. 

We also evaluate multi-source task SFT training on extractive question answering (QA), as a comparatively generous amount of multilingual data is available for this task. Specifically, we train on English data from SQuAD version 1 \citep{rajpurkar-etal-2016-squad}, all languages from MLQA \citep{lewis-etal-2020-mlqa}, and those languages from XQuAD \citep{artetxe-etal-2020-cross} which also appear in MLQA. We evaluate on the languages present in XQuAD but not in MLQA. For QA, we train for 5 epochs with batch size 12 and initial learning rate 3$e$-5.
Full details of the source languages can be found in Appendix \ref{sec:languages}.

We use an equivalent reduction factor of 1 for all tasks, following the strongest setting from our single-source experiments. Except as stated above, the training configuration and hyper-parameters are the same as for single-source training.

\section{Results and Discussion}
\label{sec:results}

\begin{table*}[t]
    \centering
    \footnotesize
        \def\arraystretch{0.99}
    {\begin{tabular}{l|l|ll|l|l}
        \toprule
        & \textbf{POS} & \multicolumn{2}{c|}{\textbf{DP}} & \textbf{NER} & \textbf{NLI} \\
        & Accuracy & UAS & LAS & F1 score & Accuracy \\
        \midrule
        \method{LT-SFT} & \textbf{71.1} (1) & \textbf{57.1} (1) & \textbf{37.8} (1) & \textbf{71.7} (1) & \textbf{51.4} (1) \\
        \method{rand-SFT} & 69.2 (1) & 54.3 (1) & 33.9 (1) & - & - \\
        \method{MAD-X} & 68.6 (16) & 54.6 (2) & 34.1 (1) & 69.9 (8) & 49.5 (2) \\
        \method{BitFit} & 58.1 & 45.7 & 23.9 & 54.9 & 38.3 \\
        \method{LT-SFT TA-only} & 51.3 (32) & 39.1 (1) & 19.9 (1) & 55.3 (8) & 39.9 (4) \\
        \method{MAD-X TA-only} & 52.1 (32) & 38.9 (1) & 19.5 (1) & 52.4 (32) & 41.7 (4) \\
        \bottomrule
    \end{tabular}}
    \caption{Results of zero-shot cross-lingual transfer evaluation averaged over all languages when best equivalent reduction factor (shown in parentheses after each result) is chosen.}
    \label{tab:best-results}
\end{table*}

\begin{figure*}[!t]
	\centering
	\begin{subfigure}{0.45\linewidth}
    	\includegraphics[width=\linewidth]{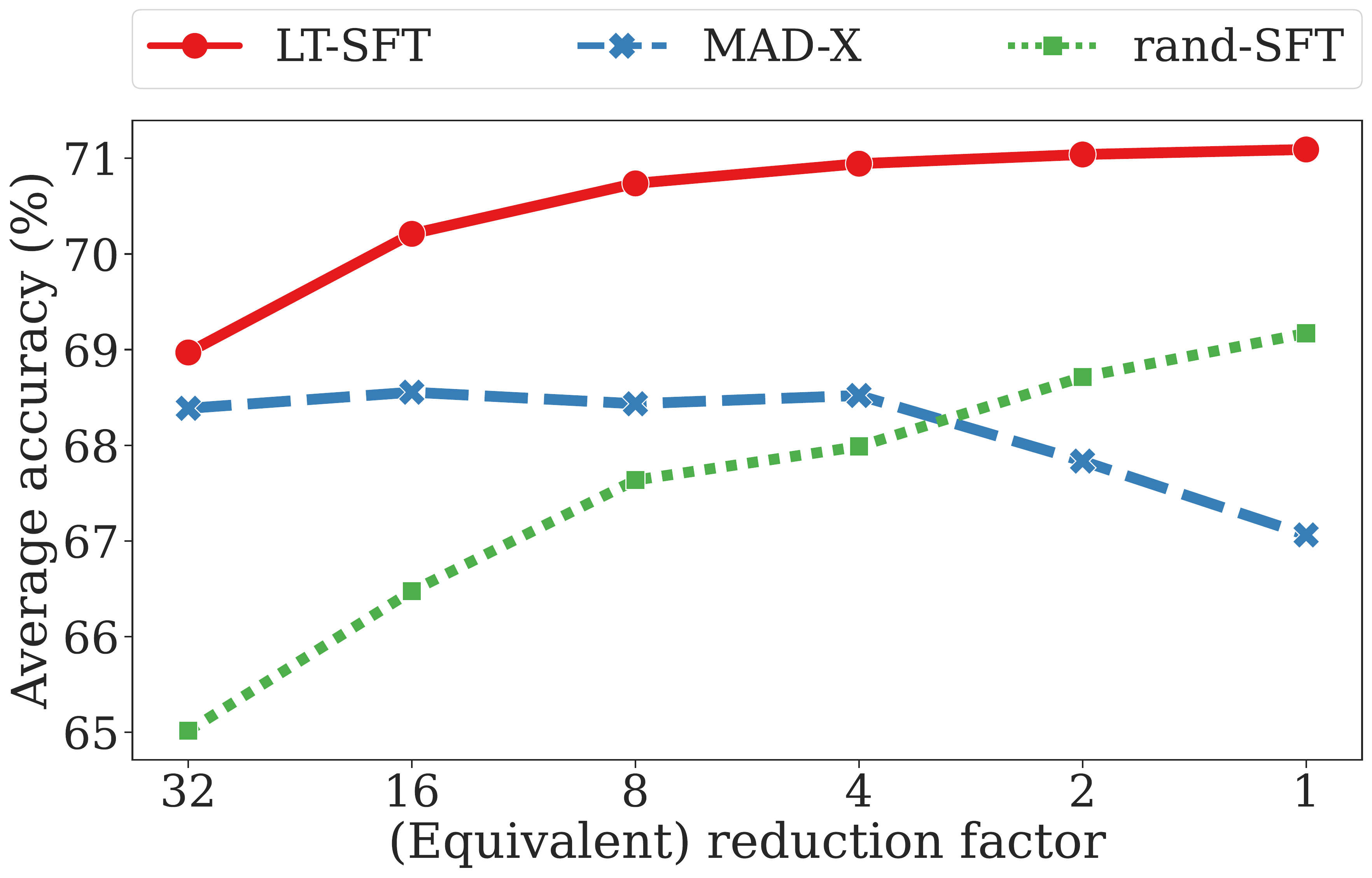}
    	\caption{Part-of-Speech Tagging}
    	\label{fig:pos_tagging}
	\end{subfigure}
	\begin{subfigure}{0.45\linewidth}
    	\includegraphics[width=\linewidth]{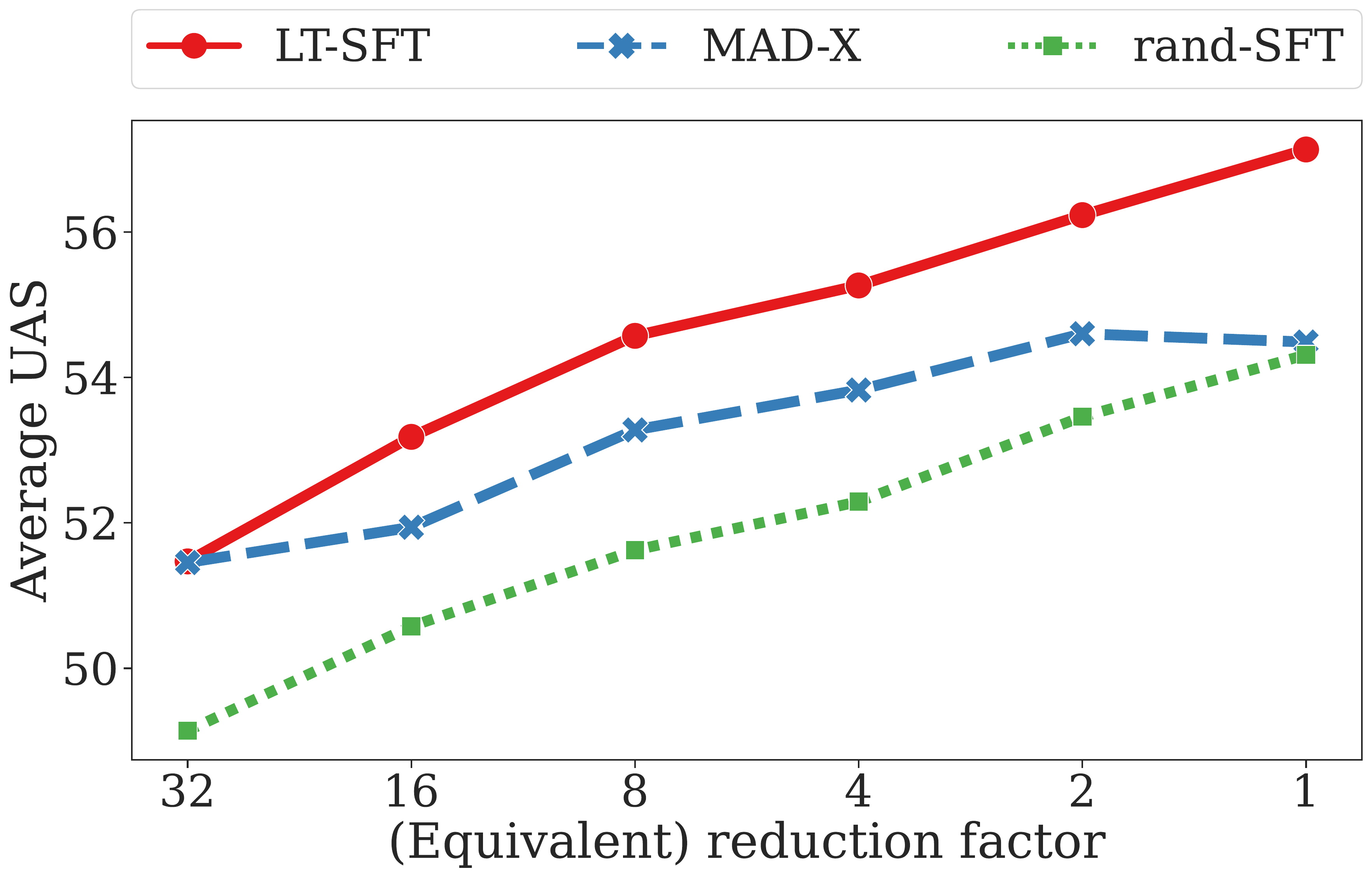}
    	\caption{Dependency Parsing (DP)}
    	\label{fig:dp}
	\end{subfigure}
	\begin{subfigure}{0.45\linewidth}
    	\includegraphics[width=\linewidth]{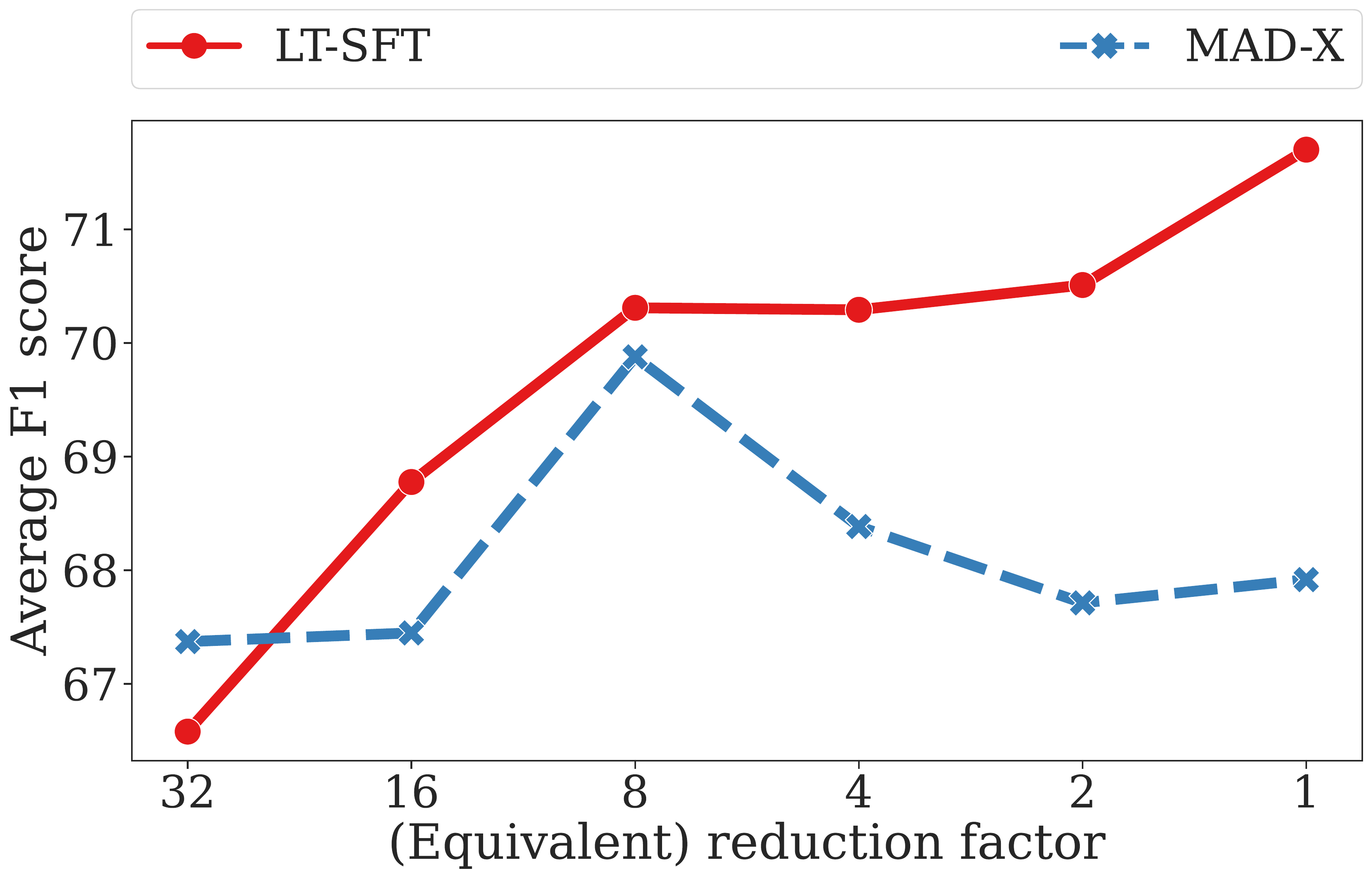}
    	\caption{Named Entity Recognition (NER)}
    	\label{fig:ner}
	\end{subfigure}
	\begin{subfigure}{0.45\linewidth}
    	\includegraphics[width=\linewidth]{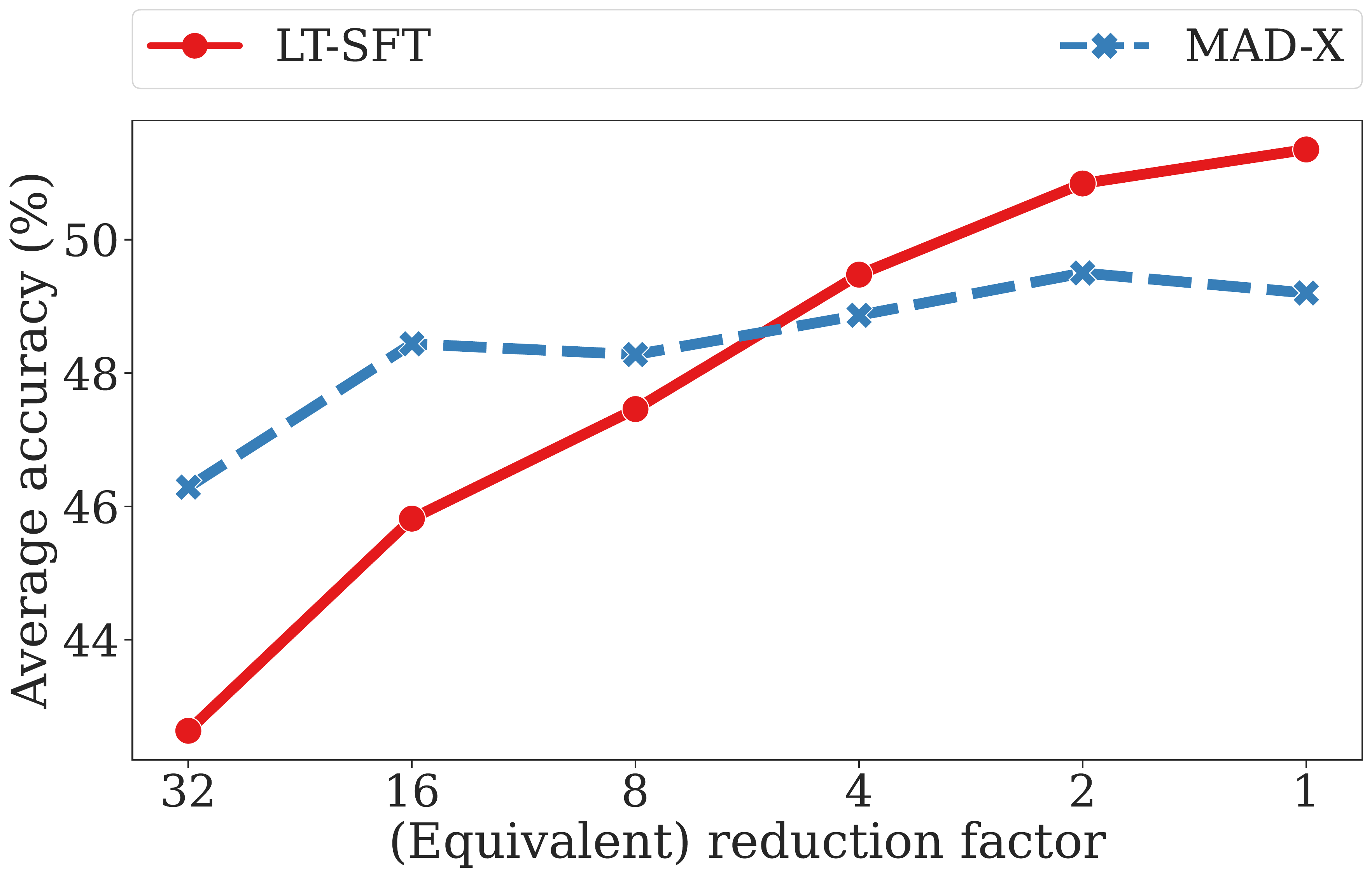}
    	\caption{Natural Language Inference (NLI)}
    	\label{fig:nli}
	\end{subfigure}
	\caption{Zero-shot cross-lingual transfer evaluation of Lottery-Ticket Sparse Fine-Tuning (\method{LT-SFT}), Random Sparse Fine-Tuning (\method{rand-SFT}), and adapter-based \method{MAD-X} over four tasks with varying numbers of trainable parameters during task adaptation. Results are averages over all target languages.}
	\label{fig:main-results}
\end{figure*}

We report the average test performance of zero-shot cross-lingual transfer for the best reduction factor (or equivalent $K$) in Table \ref{tab:best-results}. Some patterns emerge across all four tasks: first, \method{LT-SFT} consistently outperforms all the baselines. In particular, it surpasses the state-of-the-art \method{MAD-X} across all tasks, with gains of 2.5 accuracy in part-of-speech tagging, 2.5 UAS and 3.7 LAS in dependency parsing, 1.8 F1 score in named entity recognition, and 1.9 accuracy in natural language inference. Compared to \method{rand-SFT}, its superior performance demonstrates the importance of selecting ``winning tickets'' rather than a random subset of parameters. 
Secondly, the results demonstrate the importance of language SFTs/adapters for specializing pretrained models to unseen languages, as they bring about a large increase in performance across the 4 tasks compared to the corresponding settings with task adaptation only (\method{TA-only}).

We remark that LT-SFT's zero-shot performance also exceeds translation-based baselines on the AmericasNLI task, achieving an average accuracy of 51.4\%, compared with the 48.7\% of the `translate-train' baseline of \citet{ebrahimi-etal-2021-americasnli}.

In Figure \ref{fig:main-results}, we provide a more detailed overview of average cross-lingual model performance across a range of different reduction factors. The results for the \method{LT-SFT} and \method{rand-SFT} methods generally improve or stay steady as the number of trainable task parameters increases. On the contrary, there is not such a trend for \method{MAD-X}, as lower reduction factors may degrade its results. This makes it easier to choose a good setting for this hyper-parameter when using SFT. Moreover, it is worth stressing again that, contrary to \method{MAD-X}, this hyper-parameter does not affect inference time.

\begin{table*}[!t]
    \centering
    \footnotesize
    {\begin{tabular}{l|ccccc}
        \toprule
        & el & ro & ru & th & tr \\
        \midrule
        XLM-R Base, full FT & 71.1/54.3 & 78.3/63.7 & 74.1/57.8 & 67.1/55.7 & 67.5/51.1 \\
        XLM-R Large, full FT \citep{artetxe-etal-2020-cross} & 79.8/61.7 & 83.6/69.7 & 80.1/64.3 & 74.2/62.8 & \textbf{75.9}/\textbf{59.3} \\
        XLM-R Base MS, LT-SFT & \textbf{81.9}/\textbf{65.5} & \textbf{86.3}/\textbf{73.3} & \textbf{81.4}/\textbf{64.6} & \textbf{82.4}/\textbf{75.2} & 75.2/58.6 \\
        \bottomrule
    \end{tabular}}
    \caption{Results of zero-shot cross-lingual transfer evaluation on XQuAD \citep{artetxe-etal-2020-cross}, restricted to languages which do not appear in MLQA \citep{lewis-etal-2020-mlqa} (see \S\ref{ss:exp-ms}) in the format F1/exact match score. ``Full FT'' denotes full fine-tuning, MS denotes multi-source training, where additional data from MLQA and XQuAD is utilized, LT-SFT denotes Lottery-Ticket Sparse Fine-Tuning.}
    \label{tab:multisource-qa}
\end{table*}

\begin{figure*}[!t]
\centering
\begin{subfigure}[b]{0.4\textwidth}
    \centering
    \includegraphics[width=\textwidth]{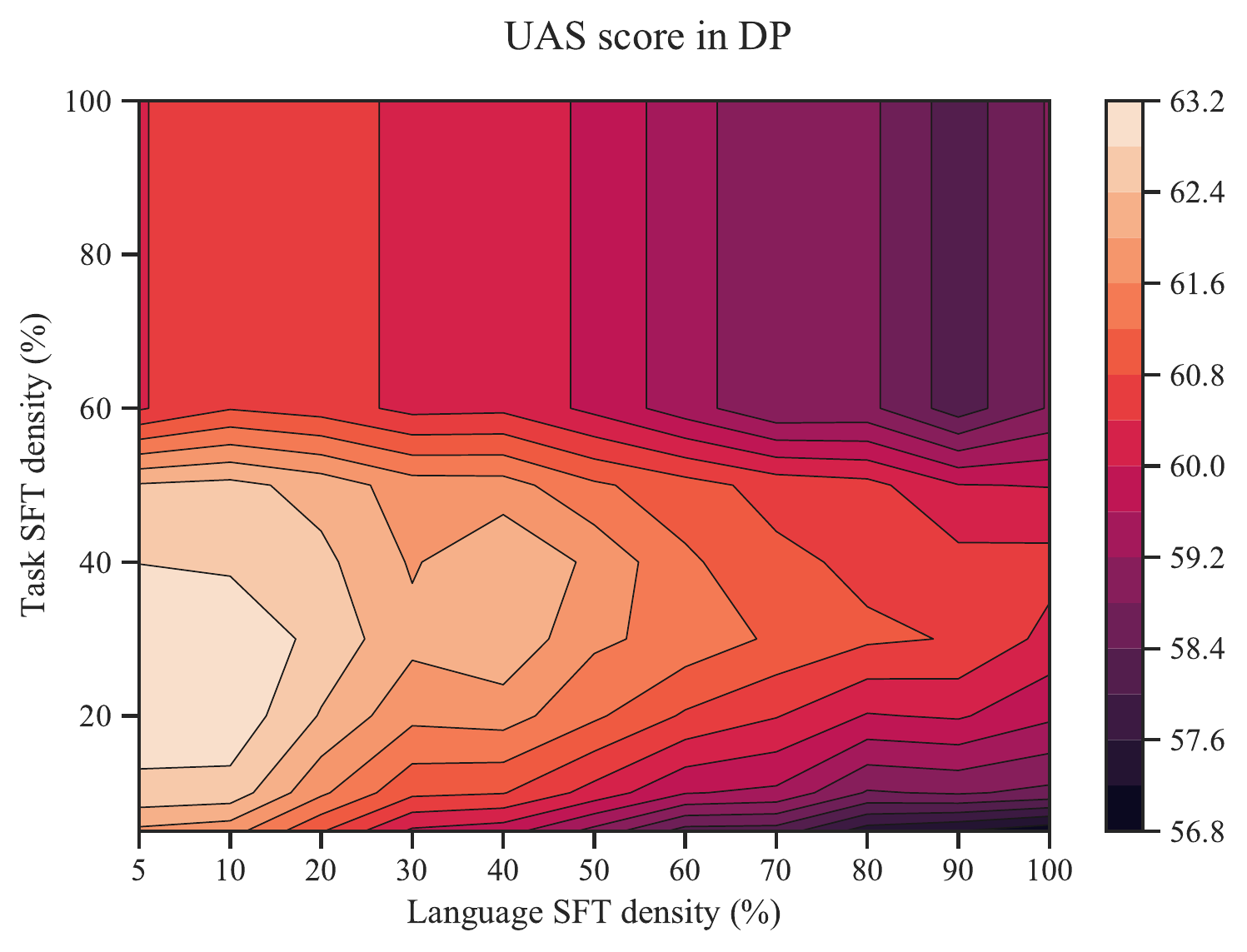}
\end{subfigure}
\begin{subfigure}[b]{0.4\textwidth}
    \centering
    \includegraphics[width=\textwidth]{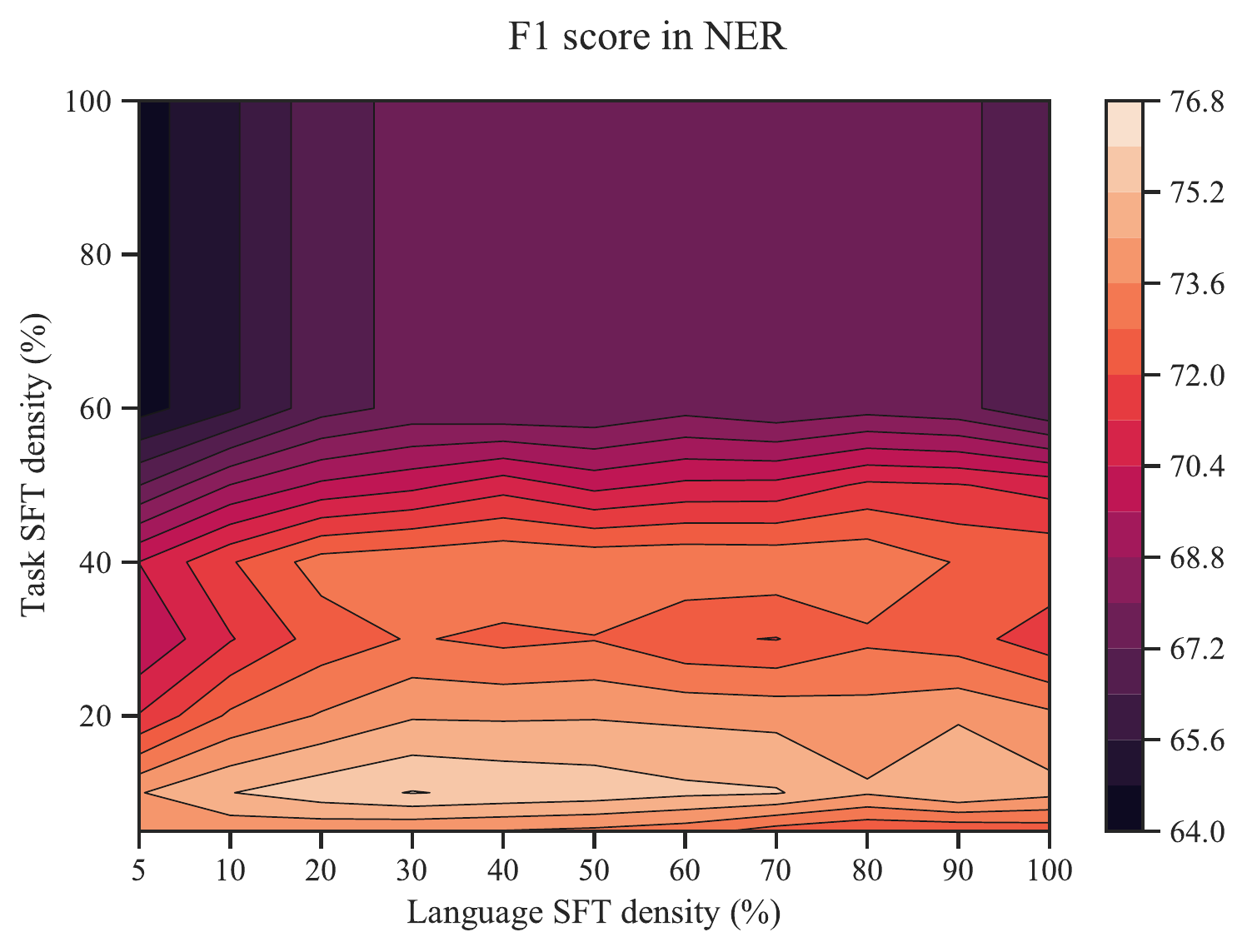}
\end{subfigure}
\caption{Performance of LT-SFT on DP and NER controlling for the sparsity of task and language fine-tuning. Results are averaged over several selected languages. Denser fine-tunings may interfere with each other and consequently degrade the model performance.}
\label{fig:interference}
\vspace{-2mm}
\end{figure*}

\method{BitFit} performs much worse than the other methods which perform language adaptation across all tasks. Bearing in mind the strong trend towards increasing performance with increasing $K$ for the other SFT methods, it seems likely that \method{BitFit}, with two orders of magnitude fewer trainable parameters, lacks the capacity to learn effective task and language SFTs.

For additional results at the level of individual languages and an analysis of the efficacy of language adaptation for high- versus low- resource target languages, we refer the reader to Appendix~\ref{sec:results-by-language}.

\subsection{Multi-Source Training}

\begin{table}[t]
    \centering
    \footnotesize
    \def\arraystretch{0.93}
    \resizebox{\linewidth}{!}
    {\begin{tabular}{l|cc|c}
        \toprule
        & DP UAS & DP LAS & NLI Accuracy \\
        \midrule
        \method{single source} & 57.1 & 37.8 & 51.4 \\
        \method{multi-source} & \textbf{64.3} & \textbf{47.6} & \textbf{53.1} \\
        \bottomrule
    \end{tabular}}
    \caption{Results of zero-shot cross-lingual transfer evaluation of single- vs. multi-source LT-SFT task training averaged over all target languages.}
    \label{tab:multisource-results}
\end{table}

As shown in Table \ref{tab:multisource-results}, multi-source LT-SFT training brings about a large improvement in zero-shot cross-lingual transfer performance on DP, and a modest improvement for NLI. This may be a result of the fact that the training set for NLI contains a relatively small number of non-English examples compared to the DP training set. Also, the AmericasNLI target languages generally have a lower degree of genealogical relatedness to the source languages compared to the DP target languages.

Table \ref{tab:multisource-qa} demonstrates that multi-source training is also beneficial to zero-shot cross-lingual transfer for QA on a series of relatively high-resource languages. In particular, LT-SFT multi-source training of XLM-R Base outperforms single-source full fine-tuning of XLM-R Large (a larger model) comfortably, and outperforms XLM-R Base single-source full fine-tuning by a significant margin. The fact that such an improvement occurs despite each of the 6 non-English source languages having more than an order of magnitude less training data than the English data from SQuAD illustrates the disproportionate advantage of multilingual source data.

\subsection{Benefits of Sparsity} \label{ssec:sparsity-analysis}
Finally, we address the following question: is sparsity responsible for preventing the interference of separate fine-tunings when they are composed? To support this hypothesis with empirical evidence, we use LT-SFT to train language\footnote{To reduce computational cost, we train language fine-tunings for a maximum of 30K steps rather than the 100K of our main experiments.} and task fine-tunings with different levels of density, i.e. the percentage of non-zero values (from 5\% to 100\%). We then evaluate all possible combinations of density levels. The results are visualized in the form of a contour plot in Figure~\ref{fig:interference} for selected combinations of tasks and languages: Buryat, Cantonese, Erzya, Maltese, and Upper Sorbian for DP, and Hausa, Igbo, Luganda, Swahili and Wolof for NER.

From Figure~\ref{fig:interference}, it emerges that the performance decreases markedly for SFTs with a density level greater than \textasciitilde30\% of fine-tuned parameters.\footnote{Note, furthermore, that levels of task fine-tuning density greater than \textasciitilde60\% do not vary in performance. This is because their subsets of parameters include embeddings of tokens never encountered during task training, which are therefore never updated even if trainable.} We speculate that this is due to the fact that sparser fine-tunings have a lower risk of overlapping with each other, thus creating interference between the different facets of knowledge they encapsulate. It must be noted, however, that alternative hypotheses could explain the performance degradation in addition to parameter overlap, such as overfitting as a result of excessive capacity. While we leave the search for conclusive evidence to future work, both of these hypotheses illustrate why enforcing sparsity in adaptation, as we propose in our method, is crucial to achieving modularity.


\section{Related Work}
\label{sec:relatedwork}


Within the framework of the Lottery Ticket Hypothesis, a series of improvements have been suggested to make the original algorithm to find winning tickets \citep{frankle2018the} more stable: after fine-tuning, \citet{frankle2019stabilizing} rewind the parameters to their values after a few iterations rather than their values before training, whereas \citet{Renda2020Comparing} also rewind the learning rate. In addition, \citet{zhou2019deconstructing} found that 1) different criteria can be used to select weights as an alternative to the magnitude of their change;
2) different rewinding methods are also effective, such as restoring the original sign, but not the value. In future work, we will investigate whether these variants also benefit our method for cross-lingual transfer, where the LTH is used for adaptation rather than pruning.





Whereas the LTH was originally conceived in the vision domain for convolutional architectures, it is also effective for pruning models trained on NLP tasks \citep{Yu2020Playing}, such as neural machine translation, and based on Transformer architectures \citep{prasanna-etal-2020-bert}. Recently, \citet{xu-etal-2021-rethinking} adapted the LTH specifically to prune pretrained models after fine-tuning.

To the best of our knowledge, \citet{NEURIPS2020_ad1f8bb9} is the only instance where winning tickets were composed in previous work. In their experiment, a set of task-specific masks were linearly combined at inference time, in order to generalize to new tasks in a continuous learning setting.

\section{Conclusion and Future Work}
We have presented a new method to fine-tune pretrained models that is both modular (like adapters) and expressive (like sparse fine-tuning). This method is based on a variant of the algorithm to find winning tickets under the framework of the Lottery Ticket Hypothesis. We infer a sparse vector of differences with respect to the original model for each individual language (by modeling unlabeled text) and each individual task (with supervised learning). The adaptations for a language and a task can then be composed with the pretrained model to enable zero-shot cross-lingual transfer. Comparing our method with the state-of-the-art baseline in several multilingual tasks, the results have indicated substantial gains across the board in both languages seen and unseen during pretraining (which includes many truly low-resource languages). 

In future work, our method offers several potential extensions. In addition to the variants to the Lottery Ticket algorithm surveyed in \S\ref{sec:relatedwork}, given the importance of sparsity for modularity (\S\ref{ssec:sparsity-analysis}), we plan to experiment with additional algorithms previously applied to pruning that can identify and fine-tune a subset of the model parameters, such as DiffPruning \citep{guo-etal-2021-parameter} and ChildTuning \citep{xu-etal-2021-raise}. Finally, given its simplicity and generality, our method is suited for many other applications of transfer learning in addition to cross-lingual transfer, such as multimodal learning, debiasing, and domain adaptation. The code and models are available online at \url{https://github.com/cambridgeltl/composable-sft}.

\section*{Acknowledgements}
{\scriptsize\euflag}  Alan wishes to thank David and Claudia Harding for their generous support via the Harding Distinguished Postgraduate Scholarship Programme. Anna and Ivan are supported by the ERC PoC Grant MultiConvAI (no. 957356) and a Huawei research donation. We would like to thank Chiara Ponti for the graphic illustration. We also thank the anonymous reviewers for their helpful suggestions.

\bibliographystyle{acl_natbib}
\bibliography{anthology,references}

\appendix

\onecolumn

\section{Algorithm of Cross-Lingual Transfer with LT-SFT}
\label{sec:ltsftalg}

\begin{algorithm}
\caption{Cross-Lingual Transfer with Lottery-Ticket Sparse Fine-Tuning}\label{alg:ltsftalg}
\begin{algorithmic}

\Function{LtSft}{$\mathcal{D}$, $\mathcal{L}$, $\bm\theta^{(0)}$, $\eta$, $K$}
    \State $\bm\theta^{(1)} \gets \bm\theta^{(0)}$
    \While{not converged}
        \State $\bm\theta^{(1)} \gets \bm\theta^{(1)} - \eta \nabla\mathcal{L}(\bm\theta^{(1)}, \mathcal{D})$
    \EndWhile
    
    \State $\mu_i \gets \begin{cases} 1 & \mathrm{if} \; \theta_i^{(1)} \in \mathop{\mathrm{argmax}}_{\theta_1, \dots, \theta_K} |\bm\theta^{(1)} - \bm\theta^{(0)} | \\ 0 & \mathrm{otherwise} \end{cases}$
    
    \State $\bm\theta^{(2)} \gets \bm\theta^{(0)}$
    \While{not converged}
        \State $\bm\theta^{(2)} \gets \bm\theta^{(2)} - \bm\mu \odot \eta \nabla\mathcal{L}(\bm\theta^{(2)}, \mathcal{D})$
    \EndWhile
    \State $\bm\phi \gets \bm\theta^{(2)} - \bm\theta^{(0)}$
    \State \Return $\bm\phi$
\EndFunction
\\
\Function{CrossLingualTransfer}{$\mathcal{D}_{\text{src}}$, $\mathcal{D}_{\text{tar}}$, $\mathcal{D}_{\text{task}}$, $\mathcal{L}_{\text{task}}$, $\bm\theta^{(0)}$, $\eta$, $K$}
    \State $\bm\phi_{\text{src}} \gets \textproc{LtSft}(\mathcal{D}_{\text{src}}, \mathcal{L}_{\text{MLM}}, \bm\theta^{(0)}, \eta, K)$
    \State $\bm\phi_{\text{task}} \gets \textproc{LtSft}(\mathcal{D}_{\text{task}}, \mathcal{L}_{\text{task}}, \bm\theta^{(0)} + \bm\phi_{\text{src}}, \eta, K)$
    \State $\bm\phi_{\text{tar}} \gets \textproc{LtSft}(\mathcal{D}_{\text{tar}}, \mathcal{L}_{\text{MLM}}, \bm\theta^{(0)}, \eta, K)$
    \State \Return $\bm\theta^{(0)} + \bm\phi_{\text{task}} + \bm\phi_{\text{tar}}$
\EndFunction

\end{algorithmic}
\end{algorithm}

\clearpage
\section{Languages}
\label{sec:languages}

\begin{table*}[!h]
    \centering
    \scriptsize
    \resizebox{\linewidth}{!}{
    \begin{tabular}{m{0.1\textwidth} | m{0.15\textwidth} | m{0.1\textwidth} | m{0.2\textwidth} | m{0.15\textwidth}| m{0.30\textwidth}}
        \toprule
        \textbf{Task} & \textbf{Language} & \textbf{ISO Code} & \textbf{Family} & \textbf{UD Treebank} & \textbf{Corpus source(s)} \\
        \midrule
        \multirow{20}{*}{Source} & Arabic$^{\dagger,\ddag}$ & ar & Afro-Asiatic, Semitic & & \multirow{20}{*}{Wikipedia} \\
        & Basque$^*$ & eu & Basque & BDT \\
        & Bulgarian$^\dagger$ & bg & Indo-European, Slavic & & \\
        & Chinese$^{\dagger,\ddag}$ & zh & Sino-Tibetan & & \\
        & Czech$^*$ & cs & Indo-European, Slavic & PDT \\
        & English$^{*,\dagger,\ddag}$, & en & Indo-European, Germanic & EWT & \\
        & Estonian$^*$ & et & Uralic, Finnic & EDT \\
        & French$^{*,\dagger}$ & fr & Indo-European, Romance & GSD \\
        & German$^{\dagger,\ddag}$ & de & Indo-European, Germanic & & \\
        & Greek$^{*,\dagger}$ & el & Indo-European, Greek & GDT \\
        & Hindi$^{*,\dagger,\ddag}$ & hi & Indo-European, Indic & HDTB \\
        & Korean$^*$ & ko & Korean & GSD \\
        & Persian$^*$ & fa & Indo-European, Iranian & PerDT \\
        & Russian$^\dagger$ & ru & Indo-European, Slavic & & \\
        & Spanish$^{\dagger,\ddag}$ & es & Indo-European, Romance & & \\
        & Swahili$^\dagger$ & swa & Niger-Congo, Bantoid & & \\
        & Thai$^\dagger$ & th & Tai-Kadai, Kam-Thai & & \\
        & Turkish$^{*,\dagger}$ & tr & Turkic, Southwestern & BOUN \\
        & Urdu$^\dagger$ & ur & Indo-European, Indic & & \\
        & Vietnamese$^{*,\ddag}$ & vi & Austro-Asiatic, Viet-Muong & VTB \\
        \midrule
        \multirow{17}{*}{POS/DP} & Arabic & ar & Afro-Asiatic, Semitic & PUD & \multirow{16}{*}{Wikipedia} \\
        & Bambara & bm & Mande & CRB & \\
        & Buryat & bxr & Mongolic & BDT & \\
        & Cantonese & yue & Sino-Tibetan & HK & \\
        & Chinese & zh & Sino-Tibetan & GSD & \\
        & Erzya & myv & Uralic, Mordvin & JR & \\
        & Faroese & fo & Indo-European, Germanic & FarPaHC & \\
        & Japanese & ja & Japanese & GSD & \\
        & Livvi & olo & Uralic, Finnic & KKPP & \\
        & Maltese & mt & Afro-Asiatic, Semitic & MUDT & \\
        & Manx & gv & Indo-European, Celtic & Cadhan & \\
        & North Sami & sme & Uralic, Sami & Giella & \\
        & Komi Zyrian & kpv & Uralic, Permic & Lattice & \\
        & Sanskrit & sa & Indo-European, Indic & UFAL & \\
        & Upper Sorbian & hsb & Indo-European, Slavic & UFAL & \\
        & Uyghur & ug & Turkic, Southeastern & UDT & \\
        \midrule
        \multirow{9}{*}{NER} & Hausa & hau & Afro-Asiatic, Chadic & \multirow{9}{*}{N/A} & Wikipedia \\
        & Igbo & ibo & Niger-Congo, Volta-Niger & & Wikipedia \\
        & Kinyarwanda & kin & Niger-Congo, Bantu & & Wikipedia \\
        & Luganda & lug & Niger-Congo, Bantu & & Wikipedia \\
        & Luo & luo & Nilo-Saharan & & \href{https://github.com/Pogayo/Luo-News-Dataset}{Luo News Dataset} \citep{adelani-etal-2021-masakhaner} \\
        & Nigerian-Pidgin & pcm & English Creole & & JW300 \citep{agic-vulic-2019-jw300} \\
        & Swahili & swa & Niger-Congo, Bantu & & Wikipedia \\
        & Wolof & wol & Niger-Congo, Senegambian & & Wikipedia \\
        & Yor\`{u}b\'{a} & yor & Niger-Congo, Volta-Niger & & Wikipedia \\
        \midrule
        \multirow{10}{*}{NLI} & Aymara & aym & Aymaran & \multirow{10}{*}{N/A} & \citet{tiedemann-2012-parallel}; Wikipedia \\
        & Ash\'aninka & cni & Arawakan & & \citet{ortega-etal-2020-overcoming,cushimariano:prel:08,mihas:anaani:11,bustamante-etal-2020-data} \\
        & Bribri & bzd & Chibchan, Talamanca & & \citet{feldman-coto-solano-2020-neural} \\
        & Guarani & gn & Tupian, Tupi-Guarani & & \citet{chiruzzo-etal-2020-development}; Wikipedia \\
        & N\'ahuatl & nah & Uto-Aztecan, Aztecan & & \citet{gutierrez-vasques-etal-2016-axolotl}; Wikipedia \\
        & Otom\'i & oto & Oto-Manguean, Otomian & & \href{https://tsunkua.elotl.mx/about/}{H\~{n}\"{a}h\~{n}u Online Corpus} \\
        & Quechua & quy & Quechuan & & \citet{agic-vulic-2019-jw300}; Wikipedia \\
        & Rar\'amuri & tar & Uto-Aztecan, Tarahumaran & & \citet{brambila-1976-diccionario} \\
        & Shipibo-Konibo & shp & Panoan & & \citet{galarreta-etal-2017-corpus,bustamante-etal-2020-data} \\
        & Wixarika & hch & Uto-Aztecan, Corachol & & \citet{mager2018probabilistic} \\
        \midrule
        \multirow{5}{*}{QA} & Greek & el & Indo-European, Greek & \multirow{5}{*}{N/A} & \multirow{5}{*}{Wikipedia} \\
        & Romanian & ro & Indo-European, Romance & & \\
        & Russian & ru & Indo-European, Slavic & & \\
        & Thai & th & Tai-Kadai, Kam-Tai & & \\
        & Turkish & tr & Turkic, Southwestern & & \\
        \bottomrule
    \end{tabular}
    }
    \caption{Details of the languages and data used for training and evaluation of SFTs and adapters. The corpora of \citet{bustamante-etal-2020-data} are available at \url{https://github.com/iapucp/multilingual-data-peru}; all other NLI corpora mentioned are available at \url{https://github.com/AmericasNLP/americasnlp2021}. $^*$ denotes source languages for multi-source DP training; $^\dagger$ denotes source languages for multi-source NLI training; $^\ddag$ denotes source languages for multi-source QA training. English is the source language in all single-source task training experiments.}
    \label{tab:languages}
\end{table*}

\clearpage
\section{Results by Language} \label{sec:results-by-language}

\begin{table*}[!h]
    
    \begin{subtable}[h]{0.44\textwidth}
        \centering
        \footnotesize
        \resizebox{\linewidth}{!}{
        \begin{tabular}{l|cccccc}
            \toprule
            & \method{LT-SFT} & \method{rand-SFT} & \method{MAD-X} & \method{BitFit} & \method{LT-SFT TA} & \method{MAD-X TA} \\
            \midrule
            ar & 68.7 & 69.3 & 70.1 & 69.8 & 70.6 & \textbf{70.8} \\
            bm & \textbf{57.0} & 55.6 & 51.0 & 41.7 & 34.2 & 37.2 \\
            bxr & \textbf{73.2} & 71.4 & 71.9 & 64.2 & 59.5 & 62.0 \\
            fo & \textbf{87.9} & 86.5 & 85.7 & 77.3 & 72.9 & 74.1 \\
            gv & \textbf{72.0} & 68.4 & 66.9 & 44.3 & 35.4 & 37.5 \\
            hsb & \textbf{83.1} & 82.4 & 81.8 & 77.2 & 69.2 & 69.6 \\
            ja & 53.9 & \textbf{54.3} & 51.1 & 53.9 & 54.1 & 51.2 \\
            kpv & \textbf{61.8} & 56.0 & 58.5 & 39.6 & 37.1 & 35.8 \\
            mt & \textbf{80.6} & 77.6 & 73.7 & 53.6 & 32.6 & 30.9 \\
            myv & \textbf{80.3} & 71.5 & 75.6 & 54.7 & 45.7 & 48.5 \\
            olo & \textbf{82.3} & 81.7 & 79.7 & 73.1 & 62.2 & 63.4 \\
            sa & \textbf{65.3} & 63.2 & 60.9 & 50.3 & 39.8 & 45.0 \\
            sme & \textbf{78.0} & 70.4 & 72.0 & 50.6 & 43.3 & 39.4 \\
            ug & 59.1 & \textbf{64.7} & 63.7 & 43.2 & 34.0 & 36.8 \\
            yue & \textbf{66.8} & 65.6 & \textbf{66.8} & 66.2 & 64.5 & 64.1 \\
            zh & 67.5 & 68.0 & 67.6 & \textbf{69.2} & 65.9 & 67.6 \\
            \midrule
            avg & \textbf{71.1} & 69.2 & 68.6 & 58.1 & 51.3 & 52.1 \\
            \bottomrule
        \end{tabular}
        }
        \caption{POS accuracy (\%)}
        \label{tab:pos-full}
    \end{subtable}
    \hfill
    \begin{subtable}[h]{0.54\textwidth}
        \centering
        \footnotesize
        \resizebox{\linewidth}{!}{
        \begin{tabular}{l|cccccc|c}
            \toprule
            & \method{LT-SFT} & \method{rand-SFT} & \method{MAD-X} & \method{BitFit} & \method{LT-SFT TA} & \method{MAD-X TA} & \method{LT-SFT MS} \\
            \midrule
            ar & \textbf{70.8}/\textbf{53.6} & 68.7/51.6 & 69.5/51.5 & 64.0/48.6 & 68.7/53.0 & 68.6/52.3 & 81.5/69.8 \\
            bm & \textbf{43.1}/\textbf{16.5} & 39.3/14.8 & 39.1/13.6 & 33.3/8.1 & 30.0/7.8 & 29.9/6.8 & 46.4/20.6 \\
            bxr & \textbf{49.2}/\textbf{25.9} & 48.3/24.1 & 48.3/24.0 & 44.9/19.7 & 40.7/17.3 & 41.0/18.0 & 60.2/35.4 \\
            fo & \textbf{68.2}/\textbf{55.5} & 65.7/53.1 & 66.3/52.5 & 57.7/43.4 & 54.3/39.8 & 53.6/38.5 & 67.2/55.6 \\
            gv & 60.0/\textbf{42.4} & 59.0/39.1 & \textbf{61.2}/37.0 & 43.3/14.7 & 28.1/5.0 & 26.4/5.4 & 66.1/52.0 \\
            hsb & \textbf{73.7}/60.5 & 72.1/58.7 & 72.1/\textbf{61.1} & 61.7/47.7 & 55.4/42.1 & 53.5/40.9 & 87.0/79.5 \\
            ja & \textbf{36.9}/\textbf{19.7} & 34.8/18.9 & 33.0/18.9 & 34.4/18.8 & 36.0/19.3 & 33.8/18.3 & 44.0/26.9 \\
            kpv & \textbf{50.5}/\textbf{27.2} & 45.1/20.7 & 47.3/22.6 & 35.8/11.3 & 24.7/7.5 & 25.4/7.1 & 57.1/35.9 \\
            mt & \textbf{74.6}/\textbf{55.4} & 68.9/48.8 & 69.4/50.8 & 51.0/25.0 & 29.2/5.7 & 28.9/5.0 & 81.0/67.9 \\
            myv & \textbf{65.9}/\textbf{45.3} & 59.8/36.3 & 59.6/35.7 & 42.2/17.2 & 32.1/11.7 & 30.3/10.4 & 73.8/57.4 \\
            olo & \textbf{66.4}/\textbf{47.8} & 64.5/43.1 & 60.9/42.0 & 52.4/29.3 & 42.2/20.0 & 42.5/18.3 & 74.9/62.4 \\
            sa & \textbf{49.5}/\textbf{25.2} & 48.9/20.8 & 46.8/19.5 & 42.8/13.9 & 32.5/8.7 & 36.0/9.9 & 62.1/39.5 \\
            sme & \textbf{58.0}/\textbf{42.1} & 49.9/29.6 & 50.6/29.0 & 31.7/10.7 & 23.2/7.0 & 22.3/6.6 & 63.4/50.7 \\
            ug & 36.4/16.7 & 37.3/15.8 & \textbf{42.1}/\textbf{19.2} & 35.3/13.5 & 21.9/7.7 & 23.5/8.4 & 56.3/35.9 \\
            yue & \textbf{51.1}/\textbf{34.0} & 48.7/31.2 & 48.8/31.8 & 44.5/27.0 & 47.4/30.0 & 47.0/29.4 & 52.1/36.3 \\
            zh & \textbf{59.8}/37.0 & 58.2/35.6 & 58.5/\textbf{37.2} & 55.9/33.7 & 58.4/36.3 & 59.1/36.9 & 55.3/35.9 \\
            \midrule
            avg & \textbf{57.1}/\textbf{37.8} & 54.3/33.9 & 54.6/34.1 & 45.7/23.9 & 39.1/19.9 & 38.9/19.5 & 64.3/47.6 \\
            \bottomrule
        \end{tabular}
        }
        \caption{DP UAS/LAS}
        \label{tab:dp-full}
    \end{subtable}
    \hfill

    \begin{subtable}[h]{0.44\textwidth}
        \centering
        \footnotesize
        \resizebox{\linewidth}{!}{
        \begin{tabular}{l|ccccc}
            \toprule
            & \method{LT-SFT} & \method{MAD-X} & \method{BitFit} & \method{LT-SFT TA} & \method{MAD-X TA} \\
            \midrule
            hau & \textbf{83.5} & 83.4 & 50.2 & 46.5 & 44.0 \\
            ibo & \textbf{76.7} & 71.7 & 57.2 & 56.8 & 54.5 \\
            kin & \textbf{67.4} & 65.3 & 56.0 & 52.9 & 50.2 \\
            lug & \textbf{67.9} & 67.0 & 50.9 & 53.8 & 53.3 \\
            luo & \textbf{54.7} & 52.2 & 35.6 & 37.7 & 33.0 \\
            pcm & \textbf{74.6} & 72.1 & 66.8 & 74.4 & 71.0 \\
            swa & \textbf{79.4} & 77.6 & 67.4 & 69.5 & 69.6 \\
            wol & \textbf{66.3} & 65.6 & 45.0 & 37.1 & 29.8 \\
            yor & \textbf{74.8} & 74.0 & 64.7 & 69.3 & 66.6 \\
            \midrule
            avg & \textbf{71.7} & 69.9 & 54.9 & 55.3 & 52.4 \\
            \bottomrule
        \end{tabular}
        }
        \caption{NER F1-score}
        \label{tab:ner-full}
    \end{subtable}
    \hfill
    \begin{subtable}[h]{0.54\textwidth}
        \centering
        \footnotesize
        \resizebox{\linewidth}{!}{
        \begin{tabular}{l|ccccc|c}
            \toprule
            & \method{LT-SFT} & \method{MAD-X} & \method{BitFit} & \method{LT-SFT TA} & \method{MAD-X TA} & \method{LT-SFT MS} \\
            \midrule
            aym & \textbf{57.9} & 51.6 & 40.8 & 38.3 & 40.7 & 59.9 \\
            bzd & \textbf{44.4} & 44.0 & 36.7 & 37.1 & 38.3 & 46.3 \\
            cni & \textbf{47.9} & 47.6 & 34.5 & 40.9 & 44.1 & 50.3 \\
            gn & \textbf{63.5} & 58.8 & 46.4 & 44.8 & 43.3 & 69.1 \\
            hch & \textbf{42.9} & 41.5 & 36.3 & 38.4 & 40.7 & 44.4 \\
            nah & 52.7 & \textbf{53.7} & 38.8 & 41.6 & 44.2 & 53.8 \\
            oto & \textbf{48.5} & 46.8 & 39.8 & 39.7 & 40.8 & 43.3 \\
            quy & \textbf{62.0} & 58.3 & 34.5 & 38.3 & 41.5 & 68.4 \\
            shp & \textbf{50.3} & 48.9 & 38.8 & 42.1 & 44.4 & 53.2 \\
            tar & 43.5 & \textbf{43.9} & 36.7 & 37.6 & 38.8 & 42.5 \\
            \midrule
            avg & \textbf{51.4} & 49.5 & 38.3 & 39.9 & 41.7 & 53.1 \\
            \bottomrule
        \end{tabular}
        }
        \caption{NLI accuracy (\%)}
        \label{tab:nli-full}
    \end{subtable}
    
    \caption{Results achieved by various zero-shot cross-lingual transfer methods across all tasks for each language. For each (method, task) pair, the (equivalent) reduction factor with the best mean score is selected as shown in Table~\ref{tab:best-results}. \method{LT-SFT MS} denotes \method{LT-SFT} with multi-source training. \textbf{Bold} denotes best-performing method per language, excluding \method{LT-SFT MS} as its larger, more diverse dataset gives it an unfair advantage.}
    \label{fig:full-results}
\end{table*}

\begin{table*}[!h]
    \centering
    \footnotesize
    {\begin{tabular}{l|cccc|cccc|ccc}
        \toprule
        & \multicolumn{4}{c|}{\textbf{POS (accuracy)}} & \multicolumn{4}{c|}{\textbf{DP (UAS)}} & \multicolumn{3}{c}{\textbf{NER (F1)}} \\
        & ar & ja & zh & avg. & ar & ja & zh & avg. & swa & yor & avg. \\
        \midrule
        \method{LT-SFT} & 68.7 & 53.9 & 67.5 & 63.4 & 70.8 & 36.9 & 59.8 & 55.9 & 79.4 & 74.8 & 77.1 \\
        \method{rand-SFT} & 69.3 & 54.3 & 68.0 & 63.9 & 68.7 & 34.8 & 58.2 & 53.9 & - & - & - \\
        \method{MAD-X} & 70.1 & 51.1 & 67.6 & 62.9 & 69.5 & 33.0 & 58.5 & 53.7 & 77.6 & 74.0 & 75.8 \\
        \method{BitFit} & 69.8 & 53.9 & 69.2 & 64.3 & 64.0 & 34.3 & 55.9 & 51.4 & 67.4 & 64.7 & 66.0 \\
        \method{LT-SFT TA-only} & 70.6 & 54.1 & 65.9 & 63.5 & 68.7 & 36.0 & 58.4 & 54.4 & 69.5 & 69.3 & 69.4 \\
        \method{MAD-X TA-only} & 70.8 & 51.2 & 67.6 & 63.2 & 68.6 & 33.8 & 59.1 & 53.8 & 69.6 & 66.6 & 68.1 \\
        \bottomrule
    \end{tabular}}
    \caption{Results for zero-shot cross-lingual transfer evaluation of the seen languages included in the POS, DP and NER evaluations. For each method/metric pair, the best equivalent reduction factor from Table \ref{tab:best-results} is used.\\ 
    Arabic, Japanese and Chinese, which were included in the POS/DP evaluation, can be considered high-resource languages; Swahili and Yor\`{u}b\'{a}, on the other hand, were included in the NER evaluation and are arguably resource-poor. In keeping with previous work, we find that language adaptation benefits seen languages less than unseen languages and---among the former---resource-rich languages less than resource-poor languages. 
    This agrees with the intuition that lower-resource languages have greater scope for improvement through language adaptation due to the fact that they receive less signal during MMT pretraining. Interestingly, \method{BitFit} performs much more competitively on the high-resource languages than low-resource and unseen languages, suggesting that its lack of capacity is more problematic for language adaptation rather than for task fine-tuning.}
    \label{tab:seen-ud-results}
\end{table*}

\clearpage
\section{MAD-X Results with AdapterHub Adapters} \label{sec:hub-eval}
\begin{figure*}[!h]
	\centering
	\begin{subfigure}{0.45\linewidth}
    	\includegraphics[width=\linewidth]{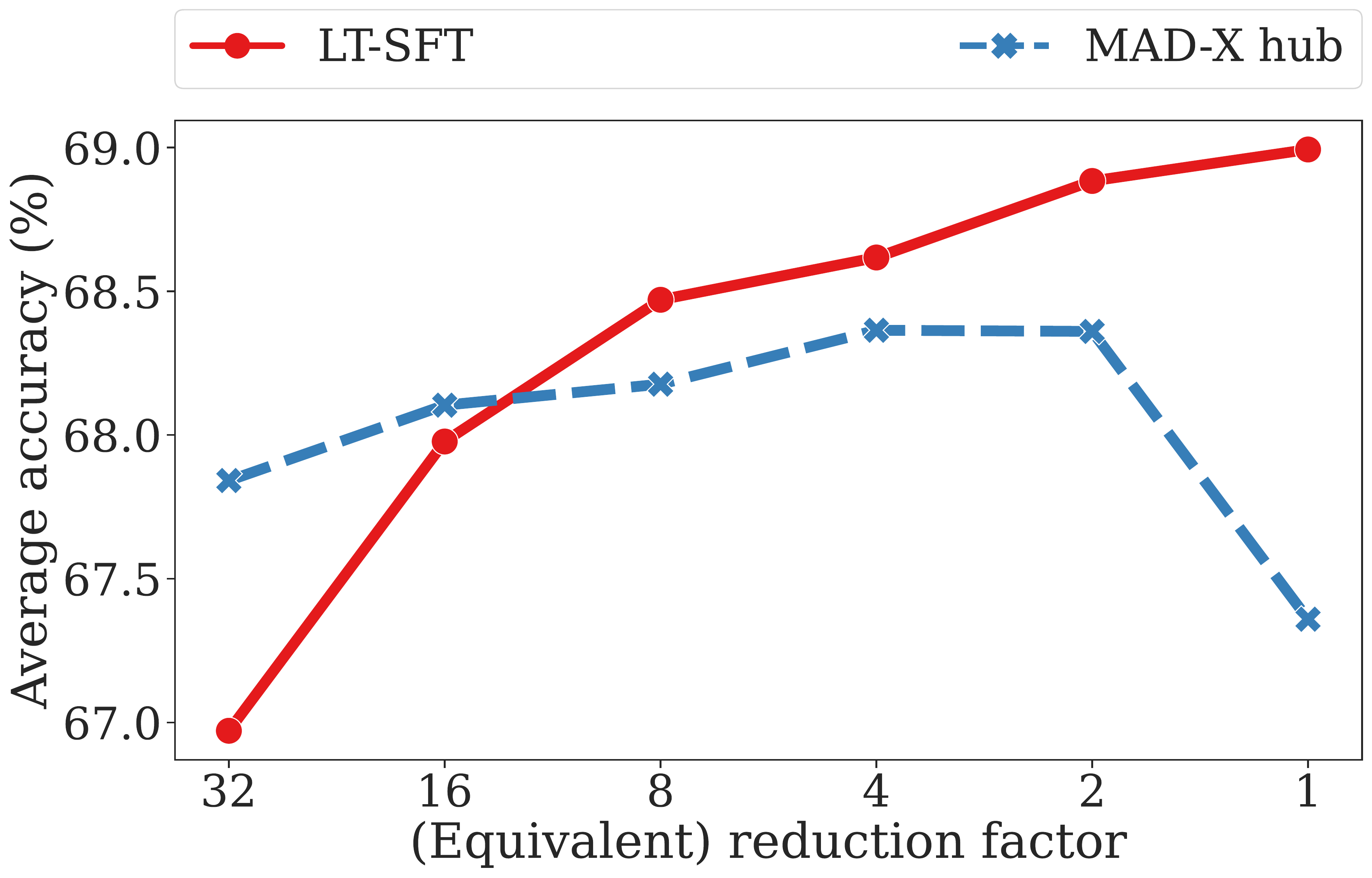}
    	\captionsetup{justification=centering}
    	\caption{Part-of-Speech Tagging\\(ar, bxr, ja, kpv, mt, myv, sme, ug, yue, zh)}
    	\label{fig:pos_tagging_hub}
	\end{subfigure}
	\begin{subfigure}{0.45\linewidth}
    	\includegraphics[width=\linewidth]{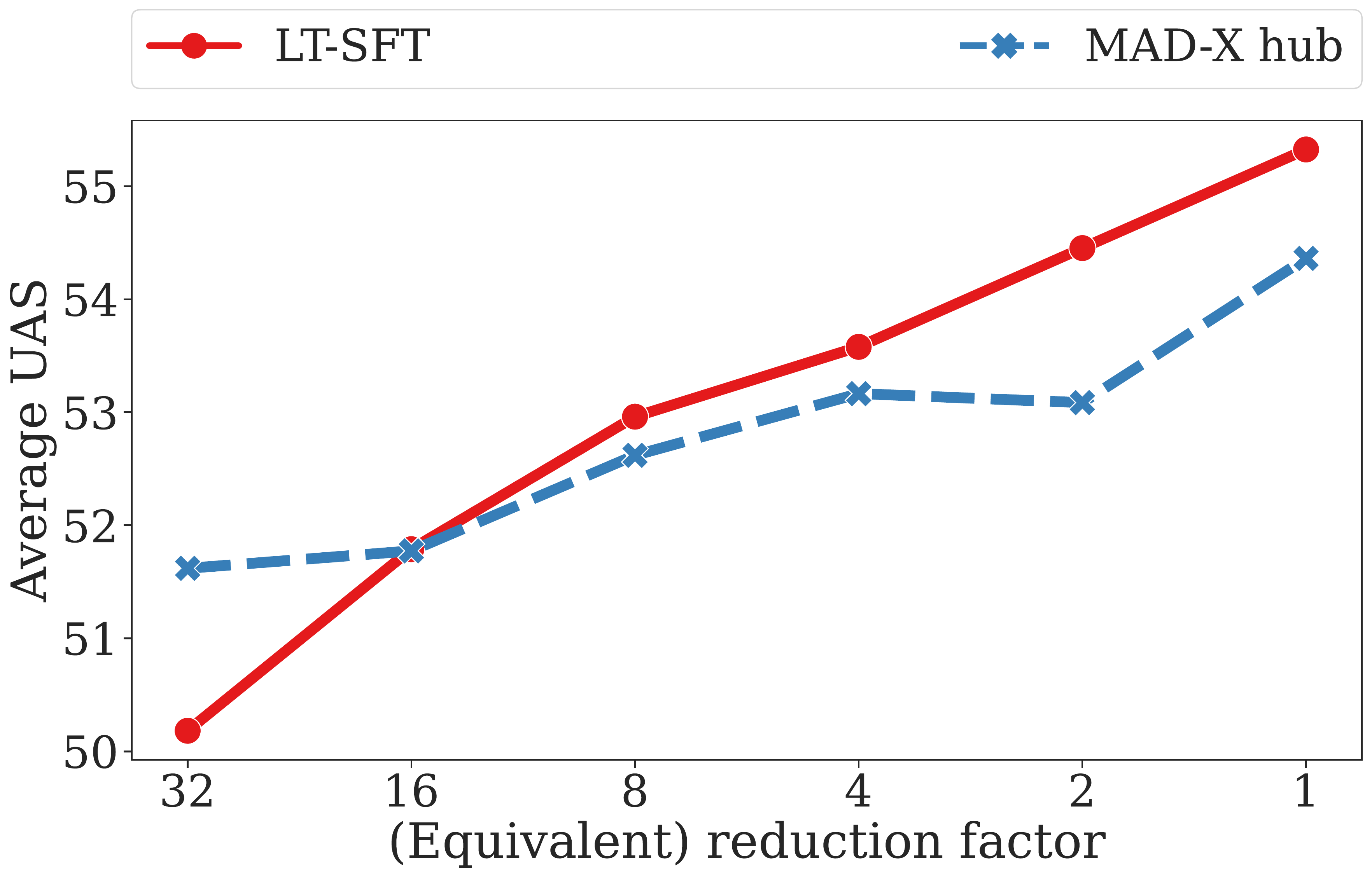}
    	\captionsetup{justification=centering}
    	\caption{Dependency Parsing (DP)\\(ar, bxr, ja, kpv, mt, myv, sme, ug, yue, zh)}
    	\label{fig:dp_hub}
	\end{subfigure}
	\begin{subfigure}{0.45\linewidth}
    	\includegraphics[width=\linewidth]{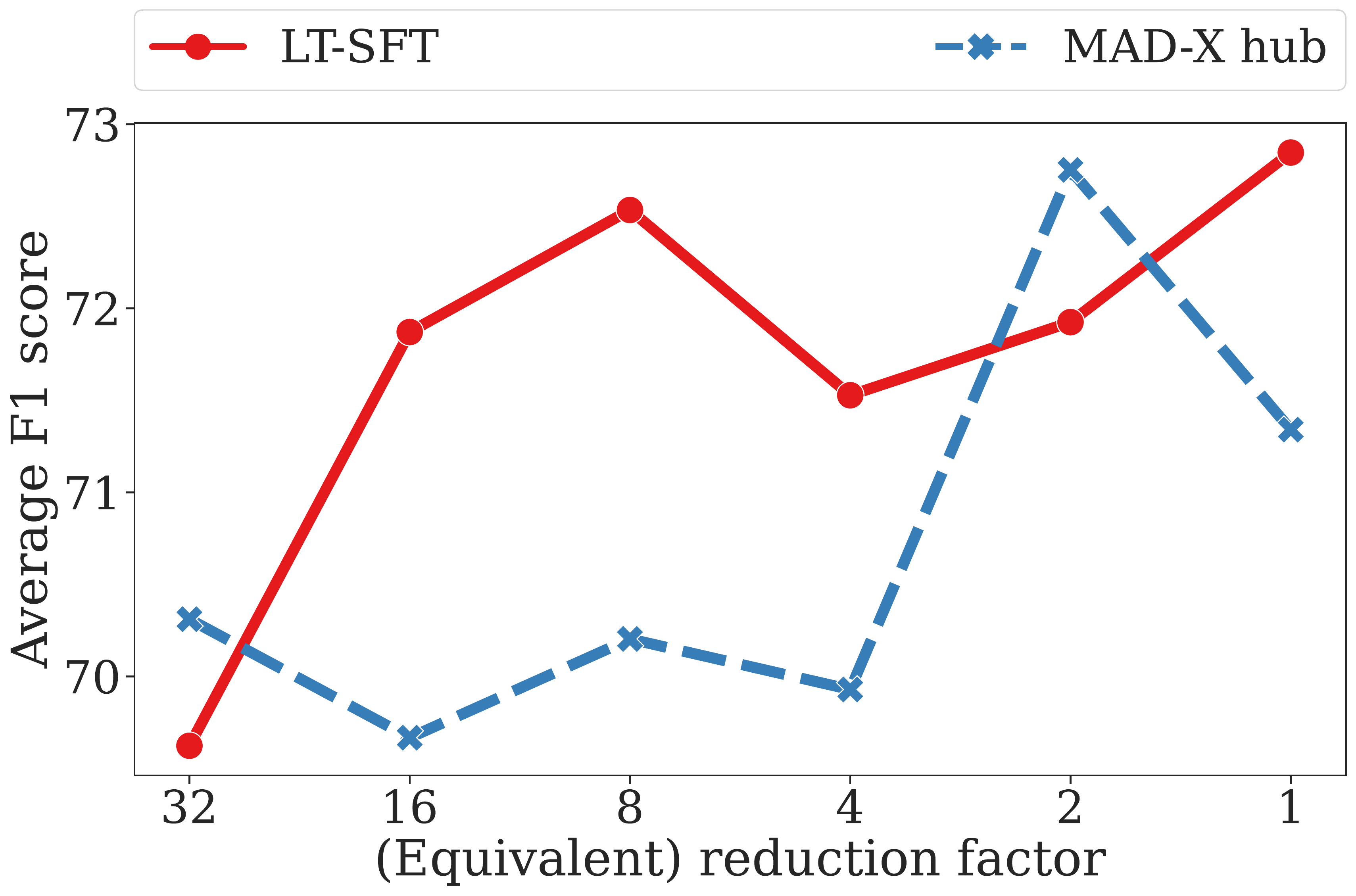}
    	\captionsetup{justification=centering}
    	\caption{Named Entity Recognition (NER) (swa, wol)}
    	\label{fig:ner_hub}
	\end{subfigure}
	\begin{subfigure}{0.45\linewidth}
    	\includegraphics[width=\linewidth]{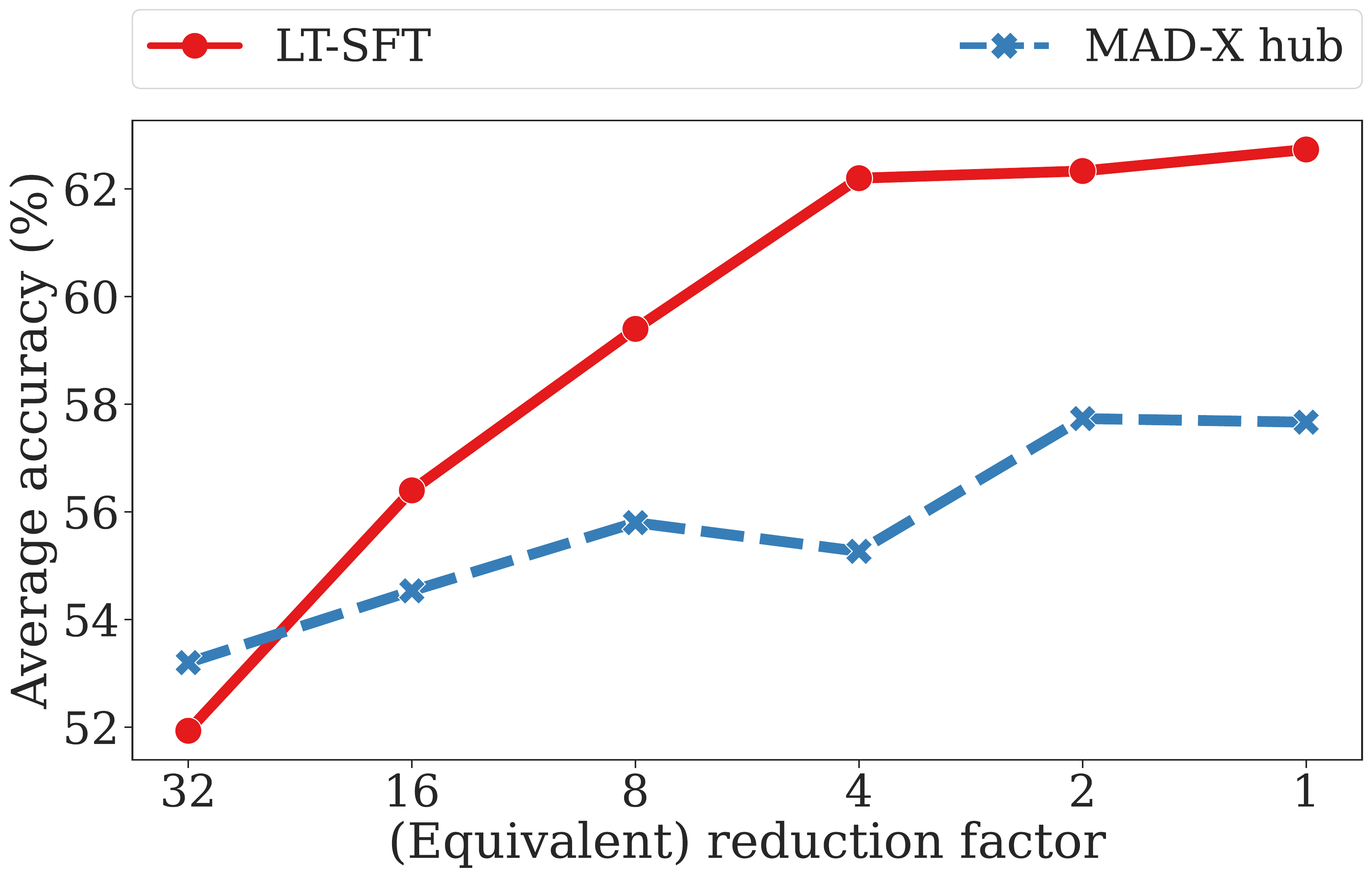}
    	\captionsetup{justification=centering}
    	\caption{Natural Language Inference (NLI) (gn, quy)}
    	\label{fig:nli_hub}
	\end{subfigure}
	\caption{Zero-shot cross-lingual transfer evaluation of Lottery-Ticket Sparse Fine-Tuning (LT-SFT) and MAD-X when pretrained language adapters from AdapterHub \citep{pfeiffer-etal-2020-adapterhub} are used during task training and evaluation. These adapters are trained for 250,000 steps with a batch size of 64, as opposed to the 100,000 steps of batch size 8 used in our experiments. LT-SFT nevertheless maintains an edge in performance across all tasks. Since AdapterHub adapters are only available for some of the languages in our evaluation, the results shown are averaged over only the languages for which they are available, indicated in the subfigure captions.}
	\label{fig:main-results-hub}
\end{figure*}

\section{Parameter Overlap between Languages}
\begin{figure}[!h]
    \centering
    \includegraphics[width=0.5\columnwidth]{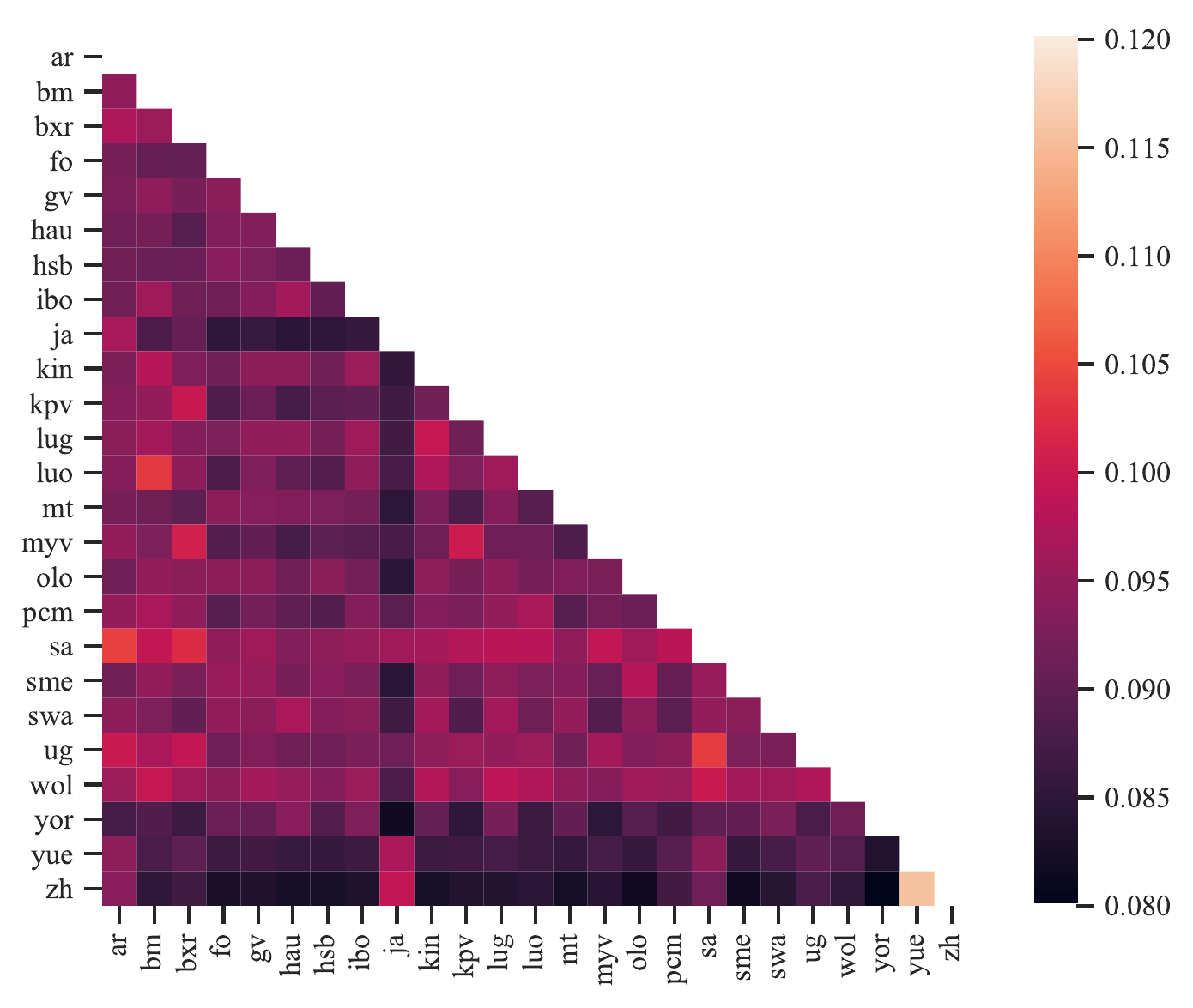}
    \caption{Percentage of parameters selected for the sparse fine-tuning of both languages in a pair.}
    \label{fig:overlap}
\end{figure}

\twocolumn
In order to understand whether similar languages also share similar sub-networks, we plot the pairwise overlap (in percentage) between parameter subsets of language SFTs in Figure~\ref{fig:overlap}. Except for a single instance (Mandarin Chinese and Cantonese) where the high overlap reflects the fact that both languages are genealogically related, we find that the overlap is small for most language pairs. The explanation, we believe, is two-fold. Firstly, most of the languages in the multilingual datasets considered in our experiments belong to separate genera and families. Therefore, a lack of correlation in parameter subsets is expected. Secondly, for a pretrained model, there exist multiple parameter subsets (``winning tickets'') with comparable performance \citep{prasanna-etal-2020-bert}. The Lottery Ticket algorithm selects randomly among these equally valid subsets. Hence, a lack of overlap does not necessarily imply the reliance on disjoint sub-networks.









\end{document}